\documentclass[ijoc,nonblindrev]{informs3} 
\usepackage{graphicx}
\usepackage{float}
\usepackage[lofdepth,lotdepth]{subfig}
\usepackage{algorithm}
\usepackage[noend]{algpseudocode}
\usepackage{tikz}
\usetikzlibrary{shapes,arrows}
\usepackage{pifont}
\usepackage[round]{natbib}
\usepackage{amssymb}
\usepackage{lineno}
\usepackage{amsmath}
\usepackage{amsfonts}
\usepackage{booktabs}
\usepackage{textcomp}
\usepackage{hyperref}
\usepackage{multirow}
\usepackage{color}
\usepackage{natbib}
\usepackage[justification=centering]{caption}

 \bibpunct[, ]{(}{)}{,}{a}{}{,}%
 \def\bibfont{\small}%
\TheoremsNumberedThrough     

\EquationsNumberedThrough    

\begin{document}



\RUNTITLE{A Hybrid Pricing and Cutting Approach for the Multi-Shift FTL Problem}

\TITLE{A Hybrid Pricing and Cutting Approach for the Multi-Shift Full Truckload Vehicle Routing Problem}

\ARTICLEAUTHORS{%
\AUTHOR{Ning Xue}
\AFF{University of Nottingham UK, \EMAIL{Ning.Xue1@nottingham.ac.uk}, \URL{}}
\AUTHOR{Ruibin Bai\footnote{Corresponding author}}
\AFF{University of Nottingham Ningbo China, \EMAIL{Ruibin.BAI@nottingham.edu.cn}, \URL{}}
\AUTHOR{Rong Qu}
\AFF{University of Nottingham UK, \EMAIL{Rong.Qu@nottingham.ac.uk}, \URL{}}
\AUTHOR{Uwe Aickelin}
\AFF{The University of Melbourne, \EMAIL{uwe.aickelin@unimelb.edu.au} \URL{}}
} 

\ABSTRACT{%
Full truckload transportation (FTL) in the form of freight containers represents one of the most important transportation modes in international trade. Due to large volume and scale, in FTL, delivery time is often less critical but cost and service quality are crucial. Therefore, efficiently solving large scale multiple shift FTL problems is becoming more and more important and requires further research. In one of our earlier studies, a set covering model and a three-stage solution method were developed for a multi-shift FTL problem. This paper extends the previous work and presents a significantly more efficient approach by hybridising pricing and cutting strategies with metaheuristics (a variable neighbourhood search and a genetic algorithm). The metaheuristics were adopted to find promising columns (vehicle routes) guided by pricing and cuts are dynamically generated to eliminate infeasible flow assignments caused by incompatible commodities. Computational experiments on real-life and artificial benchmark FTL problems showed superior performance both in terms of computational time and solution quality, when compared with previous MIP based three-stage methods and two existing metaheuristics. The proposed cutting and heuristic pricing approach can efficiently solve large scale real-life FTL problems.
}%


\maketitle

\section{Introduction}
In intermodal freight transportation, a large proportion of container transportation is carried out by barges, trains or ocean-going vessels \cite{Braekers201450}. Container movement activities between intermodal terminals, depots and shippers are also referred to as drayage operations and such activities are usually performed by trucks. Although drayage operations represent a small fraction of the total distance of an intermodal freight transportation, they constitute a substantial share of the shipping costs \cite{smilowitz2006}. Consequently, major port are facing intense competition and pressure to improve the efficiency of drayage operations.

Due to labour laws and other constraints related to the working time of drivers, shift based working schedules are becoming a common practice in the transportation industry (e.g. taxis and buses). The problem studied in this paper concerns the movement of containers (commodities) between a number of terminals (docks) within a short distance located in a large international port using a homogeneous truck fleet. The transportation time window of commodities usually spans from a few hours up to several days, and covers of multiple working shifts. Thus the problem that we are addressing is essentially different from the single-shift problem studied in most of the existing full truckload routing problems (e.g. \cite{zhang2010}, \cite{Braekers201450}) because the planning horizon covers several shifts and determining the transportation shift of each commodity forms part of the optimisation decision. 

In our earlier study \cite{Bai2015134}, a set covering model and a three-stage method were proposed for this problem. However, the computational time to optimally solve large size problems was prohibitive. \cite{chen2013task} investigate a reactive shaking variable neighbourhood search (rsVNS) and a simulated annealing hyper-heuristic method (SAHH) \cite{Chen2016} for this problem. 
The rsVNS extends the original VNS which utilises the systematic changes of multiple neighbourhood functions to achieve convergence and diversification. The SAHH applies a reinforcement learning based neighbourhood selection mechanism within a simulated annealing framework. The learning mechanism aims to adapt the algorithm to different problem instances and search scenarios by dynamically adjusting the neighbourhood selection strategies.
Both rsVNS and SAHH were able to obtain feasible but inferior solutions with less computational time compared with the three-stage method. 

The main contributions of this paper are twofold: 1) We fully explore the advantageous features of a previously proposed indirect solution encoding scheme, leading to some insightful findings of the differences between the multi-shift FTL problems and traditional pickup and delivery problems; 2) A pricing based column generation method is investigated, in conjunction with dynamic cuts. Our method is inspired by branch-price-and-cut algorithms but differs in three major ways. Firstly, we do not solve the pricing subproblem exactly. Instead, we quickly produce approximate solutions by introducing several optimisation strategies (See Section \ref{pricing-methods}). Secondly, we employ metaheuristics instead of traversing the entire search tree. Thirdly, the cuts are added after the column generation process because the infeasible flow assignments (See Section \ref{sec:dealing_with_noncompatible}) rarely occurs and they take time to be evaluated and fixed in every column generation iteration. The new algorithm can significantly speed up the solution time for large size problems. Moreover, the solution quality is also improved. Two pricing methods are proposed and tested on both real-life and artificial instances.


The remainder of this paper is organised as follows: the problem is described in detail in Section \ref{sec:problem_description}; a literature review is given in Section \ref{sec:lits} followed with the mathematical model of the problem in Section \ref{sec:model_formulation}. The proposed pricing and cutting method is illustrated in Section \ref{sec:branch-and-price} and its the computational experiments are presented in Sections \ref{sec:small_R} and \ref{sec:large_R}. Finally conclusions are drawn in Section \ref{sec:conclusion}.


\section{Problem Description}\label{sec:problem_description}
The multi-shift FTL problem is concerned with transporting a set of full truckload freights (containers) between a given number of terminals within multiple working shifts. Both the operational time windows of the freights and the planning horizon can span across several shifts. Although each container is transported in a single shift, its time window covering multi-shifts and determination of the shift in which this load is serviced (transported) forms part of the decisions to optimise. The objective is to minimise the total cost while satisfying various constraints. 

First, each full truckload commodity (container) has an available time for pickup and a deadline for delivery. Second, during each shift, a number of unit-capacity trucks start from the deport at the start of the shift, complete a number of transportation requests and then return to the depot before a shift ends. Finally, a service time is applied during both pickup and delivery. To clarify, we refer to the request of a \texttt{full truckload movement} as one \texttt{unit} of a commodity. A \texttt{commodity} is a collection of requests of full truckload freights that share identical sources, destinations and time windows. 

In the context of real-world applications that this research tries to address, the total quantities of all the requests within a planning horizon can be very large (more than 1000). The number of terminals is relatively small (less than 10) and the distances between these terminals are relatively short (all reachable within a shift). These features make this problem different from problems that are studied in previous work. It has been shown in \cite{Bai2015134} that a model based on a set covering formulation is more promising for this problem than other node based formulations. However, the features and reasons have not yet been sufficiently analysed. For the completeness of this paper, we include the model in Section \ref{sec:model_formulation}, along with a detailed discussion of its advantages and disadvantages. A literature review about the real-world applications of similar problems is given in the next section.

\section{Literature Review}\label{sec:lits}

The drayage operations problem is a typical case of bidirectional multi-shift full truckload vehicle routing problems. \cite{Bai2015134} highlight the core features of these truckload vehicle routing problems and discuss relationships with other variants of vehicle routing problems (VRP) from three aspects, including the directions of the flow, existence of consolidation or not, and length of the planning horizon. Here, we summaries the relevant research on the drayage operation problems which we broadly classify into drayage operations with and without relocation requirements of empty containers.

\subsection{Drayage problem without relocation of empty containers}
\cite{Wang200297} model a full truckload pickup and delivery problem with time windows (FT-PDPTW) as an asymmetric multiple travelling salesman problem with time windows (m-TSPTW) and propose a time-window discretisation scheme. \cite{Jula2005235} extend the m-TSPTW model with social constraints and propose an exact algorithm based on dynamic programming. Moreover, a hybrid method combining dynamic programming and genetic algorithms (GAs) is also investigated, as well as an insertion heuristic method. \cite{Chung2007252} design several types of formulations for practical container road transportation problems. The basic problem is formulated as an m-TSPTW problem, which is solved by an insertion heuristic. \cite{gendreau2015mathematical} refer to this routing problem as the one-commodity Full-Truckload Pickup-and-Delivery Problem (1-FTPDP) and present three mathematical formulations with branch-and-cut algorithms to optimally solve the model formulations.  \cite{Lai2013108} propose a new routing problem that can be viewed as a vehicle routing problem with clustered backhauls (VRPCB). Solutions are obtained with the Clarke-and-Wright algorithm and improved further by a neighbourhood based metaheuristic . This work is also compared in the study of a problem with single and double container loads \cite{ghezelsoflu2018set}. The distribution of more-than-one container per truck by different types of trucks has also been studied in \cite{vidovic2017generalized} and \cite{funke2016model}. \cite{soares2019multiple} study an FTL problem with multiple types of vehicle synchronisations. A MIP model and a heuristic solution method based on the fix-and-optimise principles are proposed.

\subsection{Drayage problems with relocation of empty containers}
Efforts to combine the planning of loaded and empty container transports are made by several authors. \cite{Coslovich2006776} analyse a fleet management problem for a container transportation company by decomposing the problem into three subproblems, which are then solved using a Lagrangian relaxation. \cite{Ileri2006} present a column generation based approach for solving a daily drayage problem. \cite{smilowitz2006} model a drayage operation with empty repositioning choices as a multi-resource routing problem (MRRP) with flexible tasks. The solution approach is a column generation algorithm embedded in a branch-and-bound framework. \cite{Imai200787}  formulate a container transportation problem as a vehicle routing problem with full container loads (VRPFC) and solve it with a subgradient heuristic based on Lagrangian relaxation. \cite{Caris20092763} extend this work and model the problem as a FT-PDPTW. A local search heuristic is proposed. The work is further extended by using a deterministic annealing algorithm suggested in \cite{caris2010}. \cite{zhang2010} improve the time window partitioning scheme used in \cite{Wang200297} for container transportation in a local area with multiple depots and multiple terminals. The results indicate that good performance can be achieved compared with a reactive tabu search (RTS) method demonstrated in \cite{Zhang2009904}. \cite{Zhang2011351} also investigate the single depot and terminal problem. Again, an RTS is proposed. \cite{Vidovic2011335} extend the problem analysed by \cite{zhang2010} and \cite{Imai200787} to the multi-commodity case and formulate it as a multiple matching problem. Solutions are obtained via a heuristic approach based on calculating utilities of matching nodes. \cite{Nossack2013666} present a new formulation for the truck scheduling problem based on a FT-PDPTW and propose a two-stage heuristic solution approach. \cite{braekers2013} investigate a sequential and an integrated approach to plan loaded and empty container drayage operations. A single- and a two-phase deterministic annealing algorithm are presented. This solution approach is further adapted in \cite{Braekers2011344} to take a bi-objective optimisation function into account. The algorithms are further improved in \cite{Braekers201450}. More recently, \cite{xie2017empty} investigate the empty container delivery problem in an intermodal transport system composed of a sea liner firm and a rail firm. Apart from transportation cost, the difference in marginal profits between the seaport and dry port is also considered in the objective function.


Some researchers examine drayage operations problems in dynamical situations. A survey on dynamic and stochastic vehicle routing problems can be found in \cite{ritzinger2016survey}.

Most of the aforementioned research work has been trying to formulate the drayage problem as some forms of classical vehicle routing problems in order to exploit the time constraint structures to prune the search space. However, as discussed in \cite{Bai2015134}, this type of formulations does not work well for problems where time related constraints are not very tight and node-based solution representations generally lead to unnecessarily large search space, resulting to inefficient solution methods. 

\subsection{Hybridising exact methods and (meta)-heuristics}
This paper studies an indirect solution representation for the multi-shift FTL problem that addresses these issues and contributes to the body of research with an efficient column generation method. In many vehicle routing applications solved by column generation, the subproblem is usually viewed as an elementary shortest path problem with resource constraints or one of its variants. Nowadays, an increasing number of hybridisations between heuristics and exact approaches are developed. These methods can provide a good compromise between solution quality and computational time as they adopt the advantages of both types of methods.
\cite{puchinger2005combining} classified hybridisation of exact algorithms and (meta)-heuristics into four types, we briefly introduce them and give examples for each: 

\textbf{1) Collaborative Combinations - sequential execution}: In this type of hybridisation, either the heuristic is executed before the exact method, or vice-versa. For example, when solving a set covering problem, a heuristic is used to generate a set of feasible columns and the exact method is used to find an optimal solution from the feasible columns. Examples of this type of hybridisation can be found in \cite{clements1997heuristic} and \cite{vasquez2001hybrid}.

\textbf{2) Collaborative Combinations - parallel or intertwined execution}: Instead of executing either heuristics or exact methods sequentially, this type of method implements the algorithms in a parallel or intertwined way. Clusters or multi-processors are used to deploy the parallel implementations. There are several frameworks proposed to facilitate such implementations, such as \cite{alba2002mallba} \cite{vidal2014unified} and \cite{lahrichi2015integrative}.

\textbf{3) Integrative Combinations - incorporating exact algorithms in heuristics}: Where exact algorithms are subordinately embedded within heuristics. For example, the solution of LP-relaxation and its dual values can be utilised in heuristically guiding neighbourhood search. Applications can be found in \cite{marino1999improving} and \cite{puchinger2004solving}.

\textbf{4) Integrative Combinations - incorporating heuristics in exact algorithms}: This type of hybridisation is analogous with the previous one, but heuristics are embedded within exact algorithms. For example, heuristics can be used to determine bounds in branch-and-bound algorithms. Heuristics can also be used to search for columns with negative costs in the branch-and-price approach. Applications of this hybridisation method can be found in \cite{puchinger2004evolutionary} and \cite{strandmark2020first}. The column generation based method proposed in this paper falls into this category.

Please refer to \cite{blum2011hybrid} and \cite{muthuraman2017comprehensive} for more comprehensive reviews of the hybridisation approach.


\section{Model Formulation}\label{sec:model_formulation}
The problem studied here can be defined on a graph $G=(N,A)$ where each node $i \in N$ represents a physical terminal (including the depot, $i=0$). An arc $(i,j)$ between nodes $i,j\in N$ is included in the arc set $A$ if the visit of $j$ can be performed immediately after $i$. A service time $t_i$ is applied to each node $i$ to represent the loading/unloading times of truckload commodities and the travel time of arc $(i,j)$ is denoted as $\mu_{ij}$. All trucks must depart from and return to node 0 (depot). Let $R$ be the set of all feasible routes that a truck can execute in a working shift without the complication of time window requirements from commodities. Therefore, each route $r\in R$ is called \texttt{distance-wise feasible}.

For a given shift $s$, the $i$-th node in route $r \in R$ (denoted as $r^i$) can only be visited within a time window $(e_{r^i}^s, l_{r^i}^s)$ where  $e_{r^i}^s$ is the earliest time that a truck covering route $r$ can start a pickup or delivery operation  while $l_{r^i}^s$ is the latest time that a truck must depart from the node. Let $t_{r^{i}}$ be the service time at node $r^i$. In each route $r$ encoding, a duplicated node is inserted if the node involves both a loading and an unloading operations (i.e. this node is both the destination and source of two different commodity flows). Therefore, if the nodes is 0-indexed in a route, loading operations are always conducted at the odd indexed nodes of a route (see Eq. (\ref{eq:2})) and unloading operations are at the even-indexed nodes.  $e_{r^i}^s$ and $l_{r^i}^s$ can be calculated using the following recursive equations:
\begin{eqnarray}
e_{r^i}^s & = & e_{r^{i-1}}^s + t_{r^{i-1}} + \mu_{r^{i-1}r^i} \qquad\forall i\in r, r\in R\label{eq:eri}\\
l_{r^i}^s & = & l_{r^{i+1}}^s - t_{r^{i+1}} - \mu_{r^{i}r^{i+1}} \qquad\forall i\in r, r\in R\label{eq:lri}
\end{eqnarray}
Let $K$ denote the set of commodities for delivery. Each commodity $k\in K$ is defined by a tuple ($o(k), d(k), Q(k), \sigma(k), \tau(k)$), which, respectively, are the origin, destination, quantity, availability time and deadline of commodity $k$. Denote $\delta_{r^i}^{ks}$ the binary variable to indicate whether the $i$-th node of route $r$ can be the loading node for commodity $k$ in shift $s$ ($\delta_{r^i}^{ks}=1$) or not ($\delta_{r^i}^{ks}=0$). To speed up the computation, $\delta_{r^i}^{ks}$ can be pre-calculated by checking the following conditions:
\begin{eqnarray}
i \texttt{ mod} \ 2 & =& 1 \label{eq:2} \\
r^i & = &o(k) \label{eq:3} \\
r^{i+1} & = &d(k) \label{eq:4}\\
l_{r^i}^s & \geq & \sigma (k) + t_{r^i} \label{eq:5} \\
e_{r^{i+1}}^s & \leq & \tau(k) \label{eq:6}
\end{eqnarray}
\begin{figure}[t]
\centering
\includegraphics[scale=0.2]{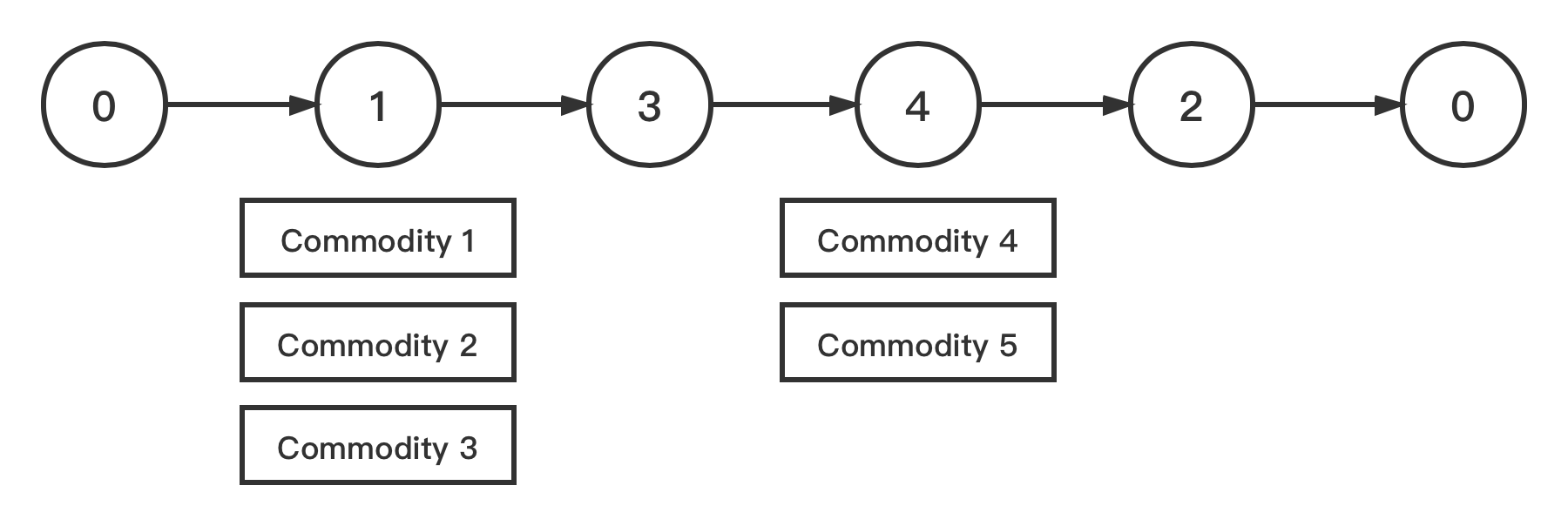}
\captionsetup{justification=centering}
\caption{Example of a routing sharing among five commodities.}
\label{fig:routeassignment}
\end{figure}

Figure \ref{fig:routeassignment} presents a simple example of a feasible truck route where 0 denotes the depot. For a 0-indexed route node list, odd numbered nodes are commodity loading nodes while even numbered nodes are unloading nodes. If a node on a route is involved with both loading and unloading, a copy of it is created so that the above rules are maintained (see more details in \cite{Bai2015134}). In this example, a truck departs from the depot and picks up a unit of commodity from either commodity 1, commodity 2 or commodity 3 from node 1 and unload the commodity at node 3. Then the truck picks another unit of commodity (either commodity 4 or commodity 5) at node 4 and unload at node 2 before the truck returns to the depot.

In summary, the following notations are used for the formulation:

\begin{flushleft}
 \textbf{Sets}
\begin{itemize}
\item $N$ : Set of nodes in the transportation network.
\item $S$ : List of time-continuous shifts in the planning horizon.
\item $R$ : Set of all feasible truck routes within a shift.
\item $K$ : Set of full truckload commodities to be delivered.
\end{itemize}

\textbf{Other parameters}
\begin{itemize}
\item $d_{r}$ : The cost (distance) of route $r$. 
\item $n$ : The maximum number of trucks available for use.
\end{itemize}

\textbf{Decision variables}
\begin{itemize}
\item $x_{r^i}^{ks}$ : Commodity flow of the $i$th node of $r$ in $s$ for servicing commodity $k \in K$.
\item $y_r^s$ : The number of times a given route $r \in R$ is used during shift $s \in S$ and $y_r^s \in \mathbf{N^+}$.
\end{itemize}
\end{flushleft}

The model for this multi-shift FTL problem can be formulated as the follows: 
\begin{equation}
\min \sum_s\sum_r{d_ry_r^s} \label{eqn:obj}
\end{equation}
\noindent subject to
\begin{eqnarray}
\sum_r{y_r^s} & \leq & n \quad \forall s\in S \label{con:asset}\\ 
\sum_s\sum_r\sum_i{x_{r^i}^{ks}} & = & Q(k) \quad \forall k\in K \label{con:demand}\\
\sum_k{x_{r^i}^{ks}} & \leq & y_r^s \quad \forall i\in r,\forall r\in R, \forall s\in S \label{con:cap}\\ 
x_{r^i}^{ks} & \le & \delta_{r^i}^{ks}y_r^s \quad\forall i\in r,\forall r\in R, \forall k\in K, \forall s\in S \label{con:arc-feas}\\ 
x_{r^i}^{ks} & = & \mathbb{Z^+} \quad\forall i\in r,\forall r\in R, \forall k\in K,  \forall s\in S \label{con:x-variable}\\
y_r^s & \in & \mathbb{Z^+} \quad \forall r\in R, \forall s\in S \label{con:y-variable}
\end{eqnarray}

The objective function (\ref{eqn:obj}) minimises the aggregated distance of all routes used in a solution. Constraint (\ref{con:asset}) restricts the total number of trucks being used in a solution. Constraint (\ref{con:demand}) ensures all the commodities are serviced (transported) entirely. Constraint (\ref{con:cap}) requires that each arc of a route in a shift can only be used $y_r^s$ times. Constraint (\ref{con:arc-feas}) makes sure that any positive $x_{r^i}^{ks}$ is feasible in terms of the source, destination and operation time window of commodity $k$. Since binary indicator $\delta_{r^i}^{ks}$ can be pre-calculated, this constraint can be eliminated by removing the corresponding flow variables $x_{r^i}^{ks}$ from the model when $\delta_{r^i}^{ks}$ takes value of 0. This is indeed how the model was implemented in our experiments because the resulting model is a lot smaller. Constraints (\ref{con:x-variable}) and (\ref{con:y-variable}) define the domains of the decision variables. 

\subsection{Merits of this solution encoding}\label{sec:model_advantage}
One of the most helpful benefits of this solution encoding scheme is the transformation of a previous m-TSPTW based non-linear model (e.g. the model proposed by \cite{Chen2016}) into a linear integer model, so it can be solved using various integer programming techniques. This was done through hiding nonlinear time related constraints into the generation of the shift-independent feasible truck route set. 

For some applications (e.g. FTL with a small number of terminals), pre-computing all feasible routes is possible since the time related constraints in this problem are slightly different from those in the traditional pickup and delivery problem with time windows (PDPTW). In this multi-shift FTL problem, each commodity $k$ has an operation time window $(\sigma(k), \tau(k)) $ defining its availability time and the delivery deadline. Time constraints require that both the pickup and delivery operations occur within this time window for commodity $k$. In PDPTW problems, two separate time windows are used, one for pickup and the other for delivery. Note that for non-time critical full truckload transportation, having one time window is reasonable since all the terminals (nodes) operate all the time, and having short time windows for both pickup and delivery does not make sense, although we acknowledge it is very different for express deliveries which are mostly for household customers. 

A second benefit of this solution representation is its capability to handle nonlinear cost functions. For example, the costs of routes could be a nonlinear, complex function of the distance. It also permits to include various other constraints related to drivers (e.g. maximum driving distance, time or preferred terminals). 

A third benefit of this solution representation is the reduced size of the search space compared with a commonly used m-TSPTW formulation, in which each container is modelled as a node in the graph. For a problem instance with hundreds or even thousands of truckload sized containers, the corresponding graph in m-TSPTW formulation could be prohibitively expensive to handle. However, in the real-world problem that we are concerned with, containers often arrive in large batches with same requirements (i.e. same O-D pairs and time windows). In an m-TSPTW formulation, any swaps of positions of these nodes (i.e. containers) in the TSP tours shall result in the same objective value (i.e. \texttt{many-to-one} mapping from solution encoding and objective space). This leads to a significantly larger search space with a plateau. In our proposed formulations, containers with the same property are grouped as one commodity, leading to a \texttt{one-to-one} mapping and a much smaller search space.

\subsection{Dealing with non-compatible commodities}\label{sec:dealing_with_noncompatible}
\begin{figure}[htbp]
\begin{center}
\includegraphics[scale=1]{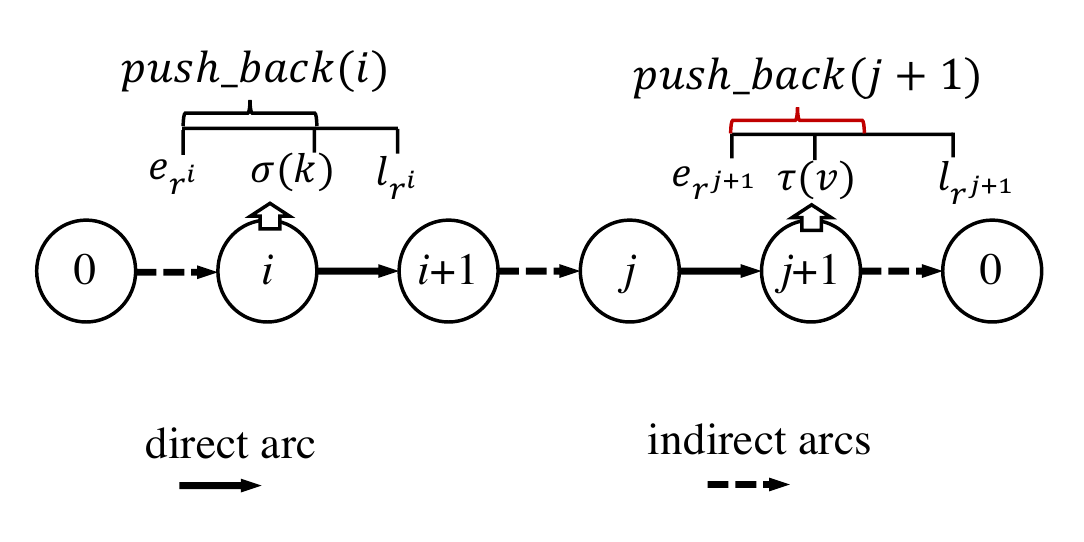}
\caption{Service time push back and non-compatible commodities.}
\label{fig:non-comptbl-commdty}
\end{center}
\end{figure}

Although for all the practical instances that we extracted from real-world problems, the FTL model in Section \ref{sec:model_formulation} produces solutions that satisfy practical constraints. However, it is possible to artificially generate problem instances that the proposed FTL model returns an infeasible solution. That is, the solution is feasible for the FTL model but may still violate the time window constraints for some commodities. This happens when two time \texttt{non-compatible} commodities are assigned to a same route and same shift. An example of such cases is illustrated in Figure \ref{fig:non-comptbl-commdty}. 

In this example, we included two non-compatible commodities ($k$ and $v$) that are serviced using a feasible route $r$ at arcs $(i, i+1)$ and $(j, j+1)$, respectively. The solution becomes infeasible when the service time of the first commodity $k$ was delayed because the commodity available time $\sigma(k)$ is later than node $i$'s earliest arrival time $e_{r^i}$. Because of this, the service time of commodity $k$ at node $i$ is pushed back by 

\begin{equation}
push\_back(i) = \sigma(k) - e_{r^i} \label{eq:push_bk}
\end{equation}

As a service node can serve multiple commodities (e.g. commodity 1, commodity 2, and commodity 3 served by node 1 in Figure \ref{fig:routeassignment}), if more than one commodities lead to $push\_back$, then $push\_back(i)$ is calculated as the maximum value of $puch\_backs$ of all commodities that served by node $i$. 

Unless there are larger push backs by other commodities in the remaining route nodes, the push back at node $i$ is propagated in its entirety to all the remaining nodes in the route. A violation of another following commodity $v$'s shipment deadline ($\tau(v)$) shall occur if the following condition is satisfied:

\begin{equation}
push\_back(j+1) > e_{r^{j+1}} - \tau(v)
\end{equation}

That is, if a push back caused by a previous commodity is greater than the difference between the earliest vehicle arrival time of its destination node and a commodity's deadline, the commodity assignments along this route become infeasible.

If the resulting solution sequentially assigns two non-compatible commodities, $k$ and $v$, at nodes $i$ and $j$, respectively, of a same route $r$ in the same shift $s$, then the following constraints should be added to ensure $v$ is not inserted at or after $k$ in the same route and shift. 
\begin{eqnarray}
x_{r^i}^{ks} \leq M\theta \quad \text{and} \quad 
x_{r^j}^{vs} \leq M(1-\theta) \quad i\le j\in r, \forall k\in K_r, \forall v\in V_r, \forall s\in S\label{con:qual1}
\end{eqnarray}
where $\theta$ is an auxiliary variable taking either 0 or 1 and $M$ is a large positive number. $K_r$ contains the commodities that $\sigma(k) > e_{r^i}$, $V_r$ contains the commodities that $\tau(v) < e_{r^{j+1}}$. Note that constraint (\ref{con:qual1}) also prevents cases of non-compatible commodity assignments at a same node. The process of generating the cuts through constraint (\ref{con:qual1}) to eliminate non-compatible commodity assignments is given in Algorithm \ref{algorithm-con:qual1}.

We do not want to strongly restrict the non-compatible commodities, as shown in the above example, $v$ is not simply forbidden to be served by the node, instead, it is still allowed to be served by the route as long as the compatibility constraint of $k$ and $v$ is not violated. From Algorithm \ref{algorithm-con:qual1} it can be seen that the procedure keeps tracking the push back time ($push\_back$) of each commodity in $K_r$ and maximum allowed push back time ($acceptable\_push\_back$) of each commodity in $V_r$. The cut will be added to the model only if any pairs of incompatible commodities were found. That means even if the service time of the commodities served between $k$ and $v$, if any, are pushed back, they are not restricted by constraint (\ref{con:qual1}) unless there are larger {$push\_back$} by other commodities result in a delay in shipment. 

\begin{algorithm}
\small
\caption{Valid cuts generation for eliminating infeasible flow assignments}
\label{algorithm-con:qual1}
{\fontsize{11}{12}\selectfont
\begin{algorithmic}[1]
\Require {$r \in R_\mathbf{x}$, where $R_\mathbf{x}$ is the set of routes used in the current solution and $\mathbf{x}$ is the vector of flow variables $x_{r^i}^{ks}$}
\State{$K_r = \emptyset$, $V_r = \emptyset$, $accu=0$}\Comment{$accu$ is the accumulated push back time along $r$}
\For{$i$ in $r$}
    \State{push\_back($i$) $\leftarrow$ 0}
    \If{($i$ mod 2=1)}
        \For{$k$ in $W(i)$}, where $W(i)$ is set of commodities serviced by node $i$. 
            \If{$\sigma(k) > e_{r^i}$}
                \State{$k$.push\_back $\leftarrow$ $\sigma(k) - e_{r^i}-accu$}
                \State{$K_r$.add($k$)}
                \State{push\_back($i$) $\leftarrow$ max($\sigma(k) - e_{r^i}$, push\_back($i$))}
            \EndIf
        \EndFor
    \EndIf
    \If{push\_back($i$)$>$0}
        \State{Propagate puch\_back($i$) to all the remaining nodes in $r$}
        \State{$accu+=$puch\_back($i$)}
    \EndIf
\EndFor

\For{$j$ in $r$}
    \If{($j$ mod 2=1)}
        \For{$v$ in $W(j)$}, where where $W(j)$ is set of commodities serviced by node $j$.
            \If{$\tau(v) < e_{r^{j+1}}$}
                \State{$v$.acceptable\_push\_back $\leftarrow$ $e_{r^{j+1}} - \tau(v)$}
                \State{$V_r$.add($v$)}
            \EndIf
        \EndFor
    \EndIf
\EndFor

\For{$k$ in $K_r$}
    \For{$v$ in $V_r$}
    \If {$k$.push\_back\_time $>$ $v$.acceptable\_push\_back}
        \State {$k$ and $v$ not compatible in $r$} \Comment{output constraint: $x_{r^i}^{ks} \leq M\theta \quad \text{and} \quad x_{r^j}^{vs} \leq M(1-\theta)$}
    \EndIf
    \EndFor
\EndFor
\end{algorithmic}
}
\end{algorithm}

\begin{table}[htbp]
\small
  \centering
  \caption{An example: A problem with 4 commodities}
    \begin{tabular}{ccccc}
    \toprule
    Commodity & Available & Deadline & Start node & End node \\
    \midrule
    $k_1$    & 13:40 & 15:05 & 1     & 2 \\
    $k_2$    & 8:00  & 13:10 & 1     & 2 \\
    $v_1$    & 8:00  & 16:00 & 3     & 4 \\
    $v_2$    & 9:00  & 15:50 & 3     & 4 \\
    \bottomrule
    \end{tabular}%
  \label{tab:small_problem}%
\end{table}%

Table \ref{tab:small_problem} gives a simple illustrative example how the valid constraints can be dynamically generated into the model to prune the search space. In this example, A total of 4 feasible routes are available for selection to deliver 4 commodity $k_1, k_2, v_1, v_2$. They are (0,1,2,3,4,0), (0,3,4,1,2,0), (0,1,2,0), (0,3,4,0) with distances of 79, 129, 64, 75, respectively. Without considering push backs, the optimal solution is to choose route 1 twice (i.e. $y_1^s=2$), with flows of $x_{1^1}^{{k_1}s}=1, x_{1^1}^{{k_2}s}=1, x_{1^3}^{{v_1}s}=1, x_{1^3}^{{v_2}s}=1$ because it satisfies formulas (\ref{eq:3}) to (\ref{eq:6}) and requires the least travel distance (158) to delivery all commodities. However, since $\sigma(k_1) > e_{1^1} (13:40 > 8:15$), the service of commodity $k_1$ at node 1 is pushed back to 13:40 by 325 minutes, which is propagated to all the remaining nodes in route 1 (the updated $e$ and $l$ after pushed back by $k_1$ are denoted as $e'$ and $l'$ in Table \ref{tab:push_back}). 

It can be seen that the push back at node 1 by commodity $k_1$ resulted in commodities $v_1$ or $v_2$ not being serviced according to the optimal solution due to $\tau(k_2) < e_{1^2} (13:10<15:00), \tau(v_1) < e_{1^4} (16:00<16:50), \tau(v_2) < e_{1^4} (15:50<16:50)$. Thus, $K_1$=\{$k_1$\}, $V_1$=\{$k_2, v_1, v_2$\} and 3 constraints ($x_{1^2}^{{k_1}s} \leq M\theta$ and $x_{1^2}^{{k_2}s} \leq M(1-\theta)$, $x_{1^2}^{{k_1}s} \leq M\theta$ and $x_{1^4}^{{v_1}s} \leq M(1-\theta)$, $x_{1^2}^{{k_1}s} \leq M\theta$ and $x_{1^4}^{{v_2}s} \leq M(1-\theta)$) are subsequently added to the model. The true optimal objective, after adding the valid constraints, increased to 208 by choosing route 1 to deliver $k_2, v_2$ and route 2 to deliver $v_1, k_1$ (i.e. $x_{1^1}^{{k_2}s}=1, x_{1^3}^{{v_2}s}=1, y_1^s=1$, $x_{2^1}^{{v_1}s}=1, x_{2^3}^{{k_1}s}=1, y_2^s=1$).

\begin{table}[htbp]
\small
  \centering
  \caption{An example: Time window $e$ and $l$ of nodes}
    \begin{tabular}{ccccccc}
    \toprule
    Node of route 1 & 0     & 1     & 2     & 3     & 4     & 0 \\
    Index of route 1 & 0     & 1     & 2     & 3     & 4     & 5 \\
    \midrule
    $e$     & 8:00  & 8:15  & 9:35  & 10:05 & 11:25 & 13:05 \\
    $l$     & 13:55 & 14:40 & 16:00 & 16:40 & 18:20 & 20:00 \\
    $e'$     & 8:00  & 13:40 & 15:00 & 15:30 & 16:50 & 18:30 \\
    $l'$     & 13:55 & 14:40 & 16:00 & 16:40 & 18:20 & 20:00 \\
    \midrule
    Node of route 2 & 0     & 3     & 4     & 1     & 2     & 0  \\
    Index of route 2 & 0     & 1     & 2     & 3     & 4     & 5 \\
    \midrule
    $e$     & 8:00  & 8:50  & 10:10 & 11:50 & 13:10 & 14:30 \\
    $l$     & 13:00 & 14:30 & 16:10 & 17:20 & 18:40 & 20:00 \\
    $e'$     & 8:00  & 8:50  & 10:10 & 13:40 & 15:00 & 16:20 \\
    $l'$     & 13:00 & 14:30 & 16:10 & 17:20 & 18:40 & 20:00 \\
    \midrule
    Node of route 3 & 0     & 1     & 2     & 0     &       &  \\
    Index of route 3 & 0     & 1     & 2     & 3     &       &  \\
    \midrule
    $e$     & 8:00  & 8:15  & 9:35  & 10:55 &       &  \\
    $l$     & 16:35 & 17:20 & 18:40 & 20:00 &       &  \\
    $e'$     & 8:00  & 13:40 & 15:00 & 16:20 &       &  \\
    $l'$     & 16:35 & 17:20 & 18:40 & 20:00 &       &  \\
    \midrule
    Node of route 4 & 0     & 3     & 4     & 0     &       &  \\
    Index of route 4 & 0     & 1     & 2     & 3     &       &  \\
    \midrule
    $e$     & 8:00  & 8:50  & 10:10 & 11:50 &       &  \\
    $l$     & 15:10 & 16:40 & 18:20 & 20:00 &       &  \\
    $e'$     & 8:00  & 8:50  & 10:10 & 11:50 &       &  \\
    $l'$     & 15:10 & 16:40 & 18:20 & 20:00 &       &  \\
    \bottomrule
    \multicolumn{7}{l}{$e$, $l$: before push back; $e'$, $l'$: after push back}\\
    \end{tabular}%
  \label{tab:push_back}%
\end{table}%

\subsection{Dealing with a very large set $R$}
The proposed model also has some problems. The most critical one is the size of the feasible route set $R$ which can increase exponentially with the number of nodes (or terminals). In \cite{Bai2015134}, some real-life problems have certain special features to permit some of nodes being merged, and a three-stage algorithm was able to find near optimal solutions. However, in addition to the excessive computational time of the three-stage algorithm, the method becomes invalid for problems that do not possess these features to permit node merging. 

In this paper we propose to use a column generation method to address this issue. The idea is to use the pricing information to guide the generation of promising feasible routes dynamically. 



\section{A Hybrid Column Generation Method}\label{sec:branch-and-price}
Column Generation is an effective approach for solving large scale integer programming problems (i.e. problems with large number of columns). It is a potentially very good method for the problem formulation stated in Section \ref{sec:model_formulation}, where the feasible route set $R$ is very large, leading to a model with a huge number of columns while the optimal solution uses a very small subset of it. We propose to use the column generation  approach for this problem in which the sub-problem (pricing problem) is solved to identify the variables that should enter the basis.

\subsection{The proposed solution framework}\label{sec:solu_frame}
\begin{figure}[t]
\centering
\includegraphics[scale=0.3]{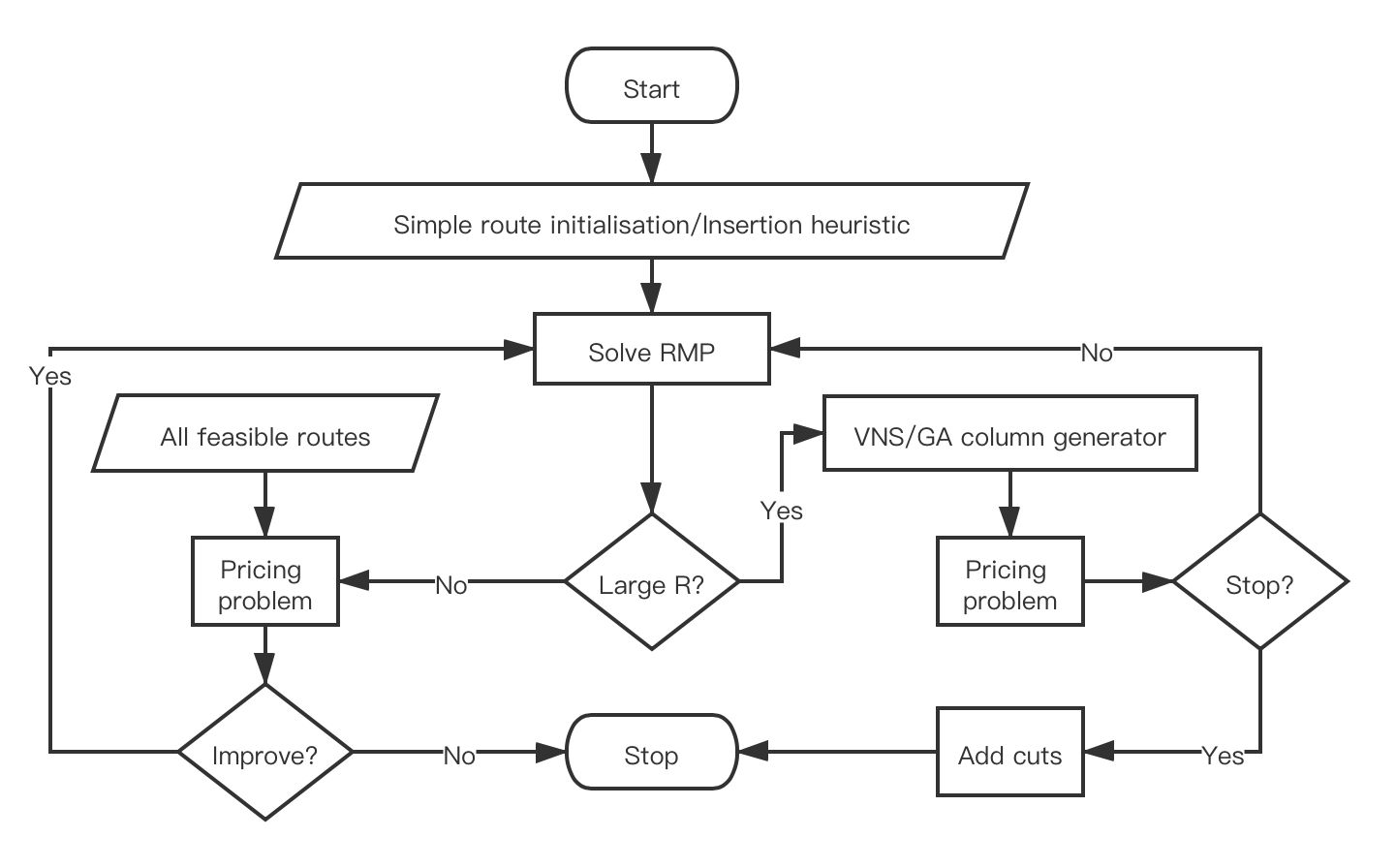}
\caption{Framework of the proposed column generation process}
\label{fig:framework}
\end{figure}

 The integer programming formulation presented in section \ref{sec:model_formulation} is also referred to as the master problem. The Restricted Master Problem (RMP) is the master problem that considers only of a subset of truck routes $R$ that are generated by the pricing problem (subproblem) using the dual information obtained from the Linear Programming Relaxation (LPR) of the RMP. The pricing problem and the LRP will be discussed in Section \ref{pricing-methods} and Section \ref{lpr}, respectively. Before the RMP is solved for the first time, no dual information is available and an initial truck routes set (see Section \ref{ini_routes}) is thus required to start the process. Then the LPR is solved to optimality and the dual information is obtained for calculating the reduced costs of routes during the pricing subproblem. The overall solution framework is outlined in Figure \ref{fig:framework}, followed by detailed steps of the procedure.

Our intial experiments showed that majority of computing time is consumed by the RMP solving. The algorithm was thus modified to accelerate convergence by adding multiple routes with negative reduced costs at each iteration. Details of the methods for the pricing problem is given in Section \ref{pricing-methods}. It is hoped that by doing this the total number of RMP calls can be reduced. This process is repeated until the stopping criteria are met. Finally, in order to obtain the integer solutions, relaxed constraints associated with $x_{r^i}^{ks}$ and $y_r^s$ are set back to their original ones during the final RMP solving.

Because the pricing problem is solved repeatedly in the column generation framework, it is crucial that the solution algorithm for the pricing subproblem is as efficient as possible. Therefore, we propose two different strategies, one for problems with small-sized $R$ and one for problems with a large $R$. For the former case, we propose to adapt an explicit enumerative generation of $R$ as \texttt{priori} and then try to solve the pricing subproblem when no column with negative reduced cost can be found. We apply a recursive algorithm to generate all feasible routes as described in \cite{Bai2015134} before the iterative procedure starts. In the case of a large $R$, we propose to use heuristic approaches (see \ref{sec:heuristic-pricing}) to solve the pricing problem and the stopping criteria of the heuristic is an limitation of the number of RMP cycles (denoted by \texttt{Finish} in Figure \ref{fig:framework}). The overall solution framework as described above is outlined in Figure \ref{fig:framework}. 

\subsection{Initial set of routes}\label{ini_routes}
Before the RMP is solved for the first time, no dual information is available and an initial set of columns is required to start the process. We apply two methods described in detail in the next two subsections to generate an initial set of columns (routes). 

\subsubsection{Simple route initialisation}\label{ini_routes_basic}
A prerequisite of constructing a basic route set is to ensure that each commodity has at least one route to service it. Thus the simplest solution is to generate a dedicated route for each commodity, in which an empty truck leaves the depot and travels to the source of a commodity, loads the commodity and delivers it to its destination. After that, the truck returns to the depot. This method works fine in some cases but may of course lead to an infeasible solution in terms of the maximum number of vehicles constraint (\ref{con:asset}). 


\subsubsection{Insertion heuristic method}\label{ini_routes_insert}
The advantage of the basic route initialisation in the previous section is its simplicity. However, very rarely will these routes be used in the optimal solutions, neither do they resemble any of the routes that are present in the optimal solutions. In this study, we proposes to use constructive heuristic methods to generate these initial routes for our column generation method. In particular, we used the same insertion heuristics described in \citep{chen2013task}. To construct routes, the task that cause minimum empty load distance is inserted by following two initialisation criteria: First, the most urgent tasks that have deadlines closer to the shift start time are inserted. The second criterion considers tasks that have earlier availability time. 

There are two benefits here. First, because the constructive heuristic produces a feasible solution for the original problem, the vehicle routes extracted from the solution shall also produce a feasible solution in our column generation method, satisfying the maximum number of vehicles constraint (constraint (\ref{con:asset})). Second, because the pricing subproblem is solved heuristically, starting from a good set of initial vehicle routes will enable the column generation method to generate high quality solutions more quickly compared to the simple route initialisation method. The proposed method will converge to a high quality solution much faster. 

\subsection{Linear programming relaxation (LPR)}\label{lpr}
Linear programming relaxation (LPR) relaxes the discrete variables constraints and differs from the model presented in Section \ref{sec:model_formulation} in the route set, which should be attained as a $R' (R' \subset R)$ resultant from the pricing subproblem instead of all feasible routes $R$. Let LPR be the relaxed model, and Let LPR-(\ref{con:asset}-\ref{con:arc-feas}) be the constraints corresponding to the constraints (\ref{con:asset}-\ref{con:arc-feas}) of the master problem.

\subsection{Pricing methods}\label{pricing-methods}
We present three different route price estimation methods: The first method obtains solution by enumerating and examining all possible commodity assignments for each route and the best route (i.e. the route with the most negative reduced cost) is selected. This method is referred to as \textbf{Pricing problem by enumeration} in this paper. For efficiency, two other pricing estimation methods (\textbf{Average pricing} and \textbf{Demand weighted average pricing}) are also investigated.\\

\noindent\textbf{Pricing problem by enumeration}. \label{sec:enumeration} Let $\alpha^s, \pi^k, \beta_{r^i}^s, \gamma_{r^i}^{ks}$ be the pricing variables for constraints from the LPR-(\ref{con:asset}-\ref{con:arc-feas}), respectively. The reduced cost for route $r$ in shift $s$ is:


\begin{equation}
    d_r+\alpha^s-\sum_{r^i}{\beta_{r^i}^s}-\sum_{r^i}\sum_k{\delta_{r^i}^{ks}\gamma_{r^i}^{ks}} \label{eq:rc}
\end{equation}

However, since routes are generated independent of shifts, the following average reduced cost is computed over all shifts for each route.

\begin{equation}
    d_r+\frac{1}{|S|}(\sum_s\alpha^s-\sum_s\sum_{r^i}{\beta_{r^i}^s}-\sum_s\sum_{r^i}\sum_k{\delta_{r^i}^{ks}\gamma_{r^i}^{ks}})
\end{equation}
where $|S|$ is the total number of shifts in the planning. Let $W=\{w_1,w_2, ..., \}$ be the set of all possible commodity assignments, each of which can be delivered by one instance of route $r$. In the example of the route in Figure \ref{fig:routeassignment}, all possible commodity assignments are $W=\{ [1,4], [1,5], [2,4], [2,5], [3,4], [3,5], ...\}$. Since there are many possible commodity assignments for a given route $r$, we evaluate them all and if the reduced cost of any given $w \in W$ is found to be negative for route $r$, it is added to the RMP. The same process is repeated for the next route $r+1$ until all feasible routes are evaluated. This searching process guarantee that the reduce cost of commodity assignment in each route is examined but its efficiency is low as some routes may contain thousands of possible commodity assignments. 

In order to obtain good results in a reasonable time, we investigated the following steps to improve efficiency: Firstly, the constraintsf LPR-(\ref{con:arc-feas}) are pre-processed offline. Given a route and its start time, the feasibility of commodity in a route can be determined by formula (\ref{eq:2})-(\ref{eq:6}) offline. This allows us to reduce a large number of decision variables that have to be handled by the model. Consequently, we lost the price values ($\gamma$) associated with the feasibility of commodity and time window of the service node.  

Secondly, we do not want to explicitly restrict which shift that a route belongs to, as the feasible route set is meant to be same across all shifts. This has benefits from management standpoints too because drivers proficiency can be improved if they are asked to follow a fixed set of routes repeatedly. Also after long run, the set of frequently used routes can become part of the knowledge system of the transportation planning and time consuming column generation procedure may not be required anymore. Therefore, the price values of arcs ($\beta$) in each shift is not used because the efficiency is substantially degraded by generating routes dependent of shifts. Fortunately, the price of an arc can be estimated by the price of all possible commodities ($\pi$) for all shifts that can be serviced by the arc. 

Thirdly, the constraint related to truck numbers is also not involved for the reduced cost calculation, due to in real-life problems, vehicle number is not critical but the efficiency is, leading to $\alpha$ taking zeros for all of our instances. In the end, we came to two approximated pricing methods, illustrated below.\\

\noindent\textbf{P1: average pricing}\label{average}
Instead of enumerating all the commodity combinations of a route and then checking the $c_r$ for each of $w_i$, a more efficient approach is to use the average prices to estimate $c_r$ approximately. More specifically, let $J$ be the set of all service nodes in $r$ (e.g. nodes $\{1,4\}$ in Figure \ref{fig:routeassignment}). Denote $V'_j$ be the set of all commodities that can be serviced by a node $j$ in $r$.  The reduced cost $c_r$ for route $r$ is calculated by the following equation:
\begin{eqnarray}
c_r = d_r-\sum_{j\in J}(\frac{1}{|V'_j|}\sum_{k\in V'_j}{\pi^k}) \label{eq:reduce_cost_avg} 
\end{eqnarray}  

\noindent\textbf{P2: demand weighted average pricing}\label{weight}
Though the commodities processed by a service node in a route share the same source and destination node, the quantity of the commodities varies from one to another. The simple average pricing method P1 fails to take into account the quantity of the commodities, so that large quantity commodities may be left unpaired to improve the efficiency. Therefore, the demand weighted average method tries to give priority to large commodities at the early stage. A weight $\omega_k$ that is proportional to the commodity quantity $Q(k)$ is used. The weighted average pricing method uses the following equation to estimate the reduced cost. 
\begin{eqnarray}
c_r = d_r - \sum_{j\in J}\sum_{k\in V'_j}{\omega_k\pi^k} \label{eq:reduce_cost_wavg}
\end{eqnarray}


\subsection{Heuristic column generator for large $R$} \label{sec:heuristic-pricing}

As can be seen from Figure \ref{fig:framework}, a heuristic column generator is used within the column generation framework. As optimally solving the pricing problem involves an expensive recursive tree search, we propose to use a variable neighbourhood search (VNS) and a genetic algorithm (GA) to tackle the pricing subproblem. The goal of the metaheuristics is to identify new columns with negative reduced costs. The idea is that, instead of generating a new column (i.e. route) from scratch, it is probably more efficient to search from the existing routes through either neighbourhood moves or route combinations (i.e. crossovers). VNS and GA are widely adopted excellent frameworks to implement these ideas. The main difference here is that the metaheuristics are guided by an objective function that heavily relies on the pricing information obtained from the linear program relaxations. 

\subsubsection{VNS}\label{sec:vns}
The pseudo-code of our VNS algorithm is given in Algorithm \ref{algorithm:vns} and the parameters of the algorithms are listed in Table \ref{algo:vns}. In our VNS method, the neighbourhood functions include \textit{swap}, \textit{2-opt}, and \textit{relocate}. These operators are very similar to those used in solving the classical VRP problems. For example, the \textit{swap} operator swaps two arcs of two different routes. The \textit{2-opt} exchanges two nodes on the same route. The \textit{relocate} operator relocates an arc from its current route to a different one. By exploring different neighbourhood structures, the method has an increased probability to detect more diversified routes than a single neighbourhood. The neighbourhood functions are called one by one in the order of \textit{swap}, \textit{2-opt} and \textit{relocate}. Once a neighbourhood function can no longer find a better set of routes, the next neighbourhood is called. If, however, a better solution (e.g. a more negative reduced cost) is found, the algorithm will restart from the first neighbourhood (i.e. \textit{swap}). 


\begin{table}[htbp]
\small
 \centering
 \caption{Abbreviations of VNS \label{algo:vns}}
 \begin{tabular}{ll}
 \hline
 $\mathbf{z}$ & current solution \\
 $R_z$ & a set of routes present in $\mathbf{z}$\\
 $i$ & index of neighbourhood \\
 $i_{max}$ & index of the last neighbourhood function \\
 $c_{min}$ & minimum reduced cost of route set \\
 $c_{min}'$ & modified minimum reduced cost of route set \\
 $maxIteration$ & max number of column generation iterations \\
 $maxColumns$ & max number of routes \\
 $columnPool$ & stores the set of best routes\\
 \hline
 \end{tabular}%
 \label{tab:algorithm_abbre}%
\end{table}%

\begin{algorithm}
\small
\caption{Pseudo-Code of VNS column generator}
\label{algorithm:vns}
{\fontsize{11}{12}\selectfont
\begin{algorithmic}[1]
\Require $\mathbf{z}$, $maxIteration$
\State $j\leftarrow 0$
\While{$j < maxIteration$}
\State $columnPool \leftarrow VNS(\mathbf{z})$ \Comment{Algorithm \ref{algorithm:vns_function}}
\State $\mathbf{z} \leftarrow RMP(columnPool)$, $j\leftarrow j+1$
\EndWhile
\State
\Return $\mathbf{z}$
\end{algorithmic}
}
\end{algorithm}

\begin{algorithm}
\small
\caption{Pseudo-Code of VNS()}
\label{algorithm:vns_function}
{\fontsize{11}{12}\selectfont
\begin{algorithmic}[1]
\Require $\mathbf{z}$, $i_{max}$, $maxColumns$
\State $i\leftarrow 1$, update $R_z$ by $\mathbf{z}$, $c_{min}\leftarrow 0$
\While{$i \leq i_{max}$}
\State $R' \leftarrow neighbourhood(R_z,i,maxColumns)$
\State $c_{min}'\leftarrow minReducedCost(R')$
\If{$c_{min}'<c_{min}$}
\State $i\leftarrow 1$, $c_{min}\leftarrow c_{min}'$
\State $columnPool \leftarrow sortByReducedCost(R',maxColumns)$
\State $columnPool \leftarrow  columnPool \cup \mathbf{z}$
\Else 
\State $i\leftarrow i+1$
\EndIf 
\EndWhile
\State
\Return $columnPool$
\end{algorithmic}
}
\end{algorithm}

Before VNS starts, the initial set of columns in $\mathbf{z}$ is generated by the insertion heuristic (\cite{chen2013task}). As shown in Algorithm \ref{algorithm:vns}, for each successive iteration, $\mathbf{z}$ is updated subsequently. Since our VNS not aims to solve the overall problem but find out a set of feasible routes with the most negative reduced costs to be solved by the RMP, the VNS based column generator is not conventionally implemented with a \textit{shaking} process. The search is guided by the pricing methods described in Section \ref{pricing-methods}. The \textit{neighbourhood($R$,$i$,$maxColumns$)} function applies the $i$-th neighbourhood function on all routes in $R_z$ to search for new feasible routes. It returns the maximum of $maxColumns$ distinct routes with negative reduced cost. The constraints related with feasible route pattern (Eq. (\ref{eq:2}) to (\ref{eq:6})) are imposed. Function $minReducedCost(R')$ returns the minimum reduced cost of route set $R'$. Function $sortByReducedCost(R_z \cup R',maxColumns)$ sorts routes in $R_z \cup R'$ by their corresponding reduced costs in an ascending order and returns the top $maxColumns$ distinct routes. The $RMP(columnPool)$ is the restricted master problem (see Section \ref{lpr}) based on the route set stored in $columnPool$ and the solution is stored in set $\mathbf{z}$.

Note that a distinctive feature of our VNS based column generation method is the joint exploitation of pricing information and the current best solution $\mathbf{z}$. While most heuristic column generation methods aim to \texttt{construct} new routes from scratch in light of new pricing information, our VNS column generation procedure (and the GA column generator) is a \texttt{perturbative} based neighbourhood search starting from existing columns in the basis. Consequently, we believe our column generation methods converge much faster than the constructive methods used in the literature.

\subsubsection{Genetic algorithm}\label{sec:ga}
\begin{figure}[htbp]
\centering
\subfloat[][Example of crossover 1]{
\includegraphics[width=0.45\textwidth]{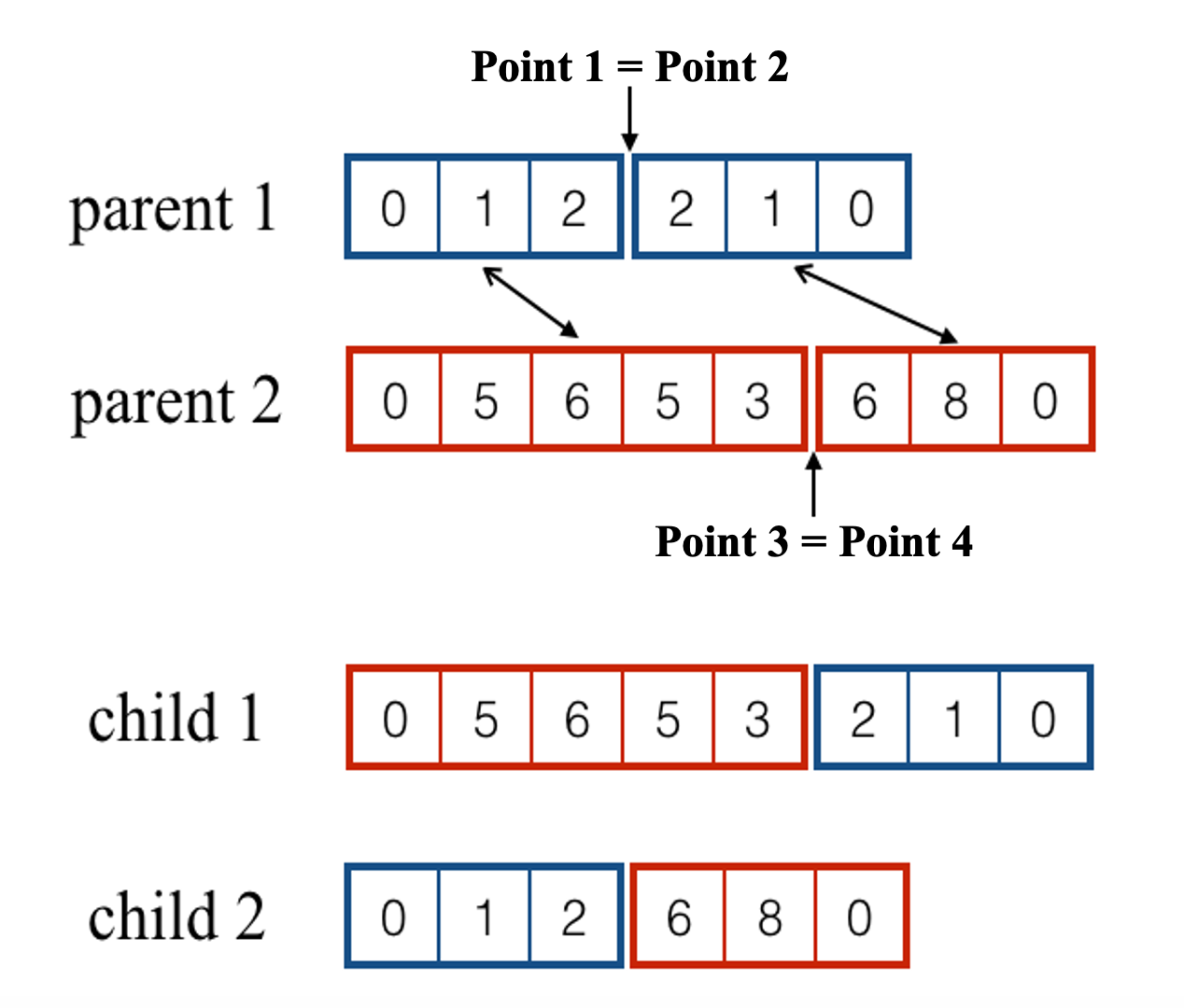}
\label{fig:crossover1}}
\subfloat[][Example of crossover 2]{
\includegraphics[width=0.49\textwidth]{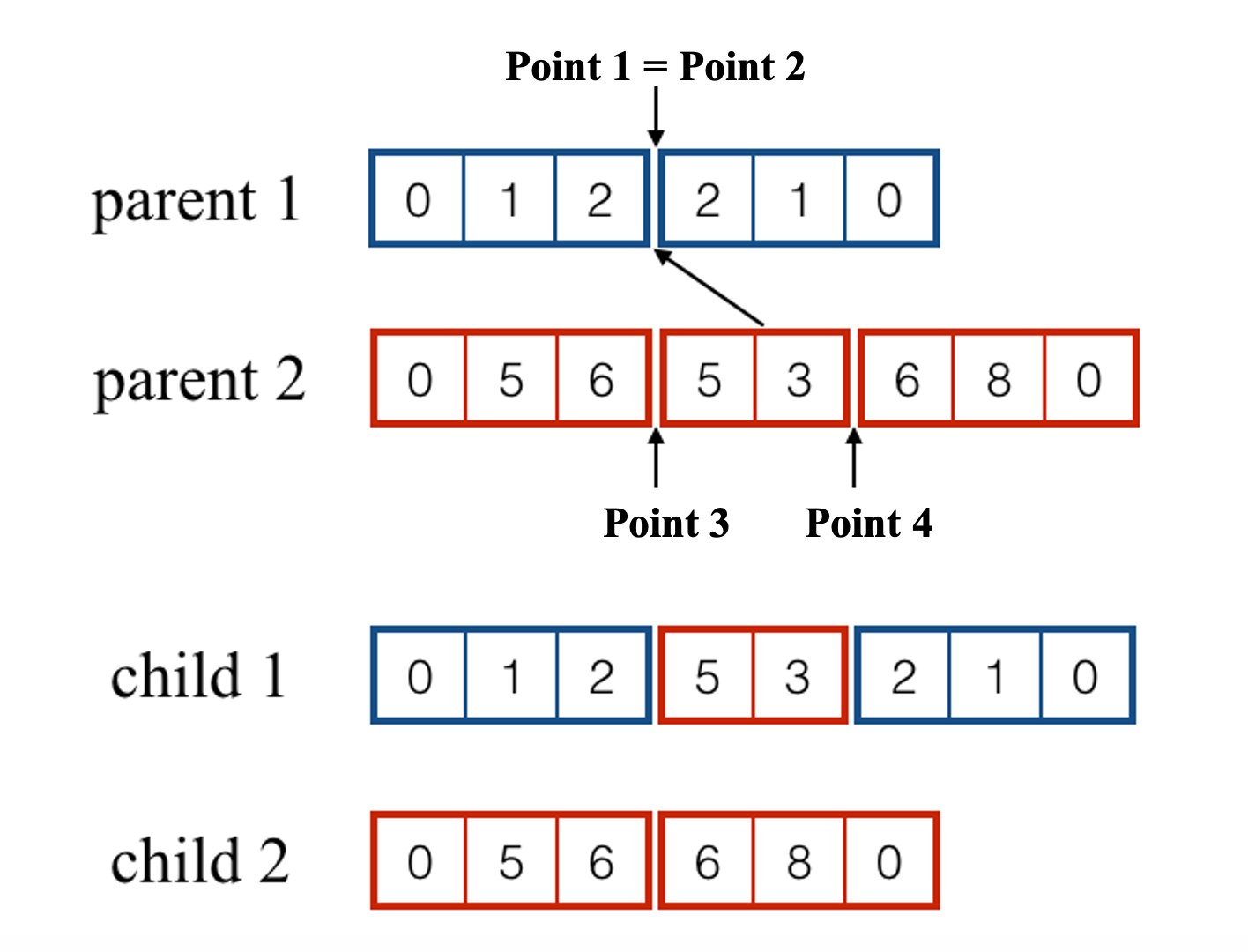}
\label{fig:crossover2}}
\qquad
\subfloat[][Example of crossover 3]{
\includegraphics[width=0.49\textwidth]{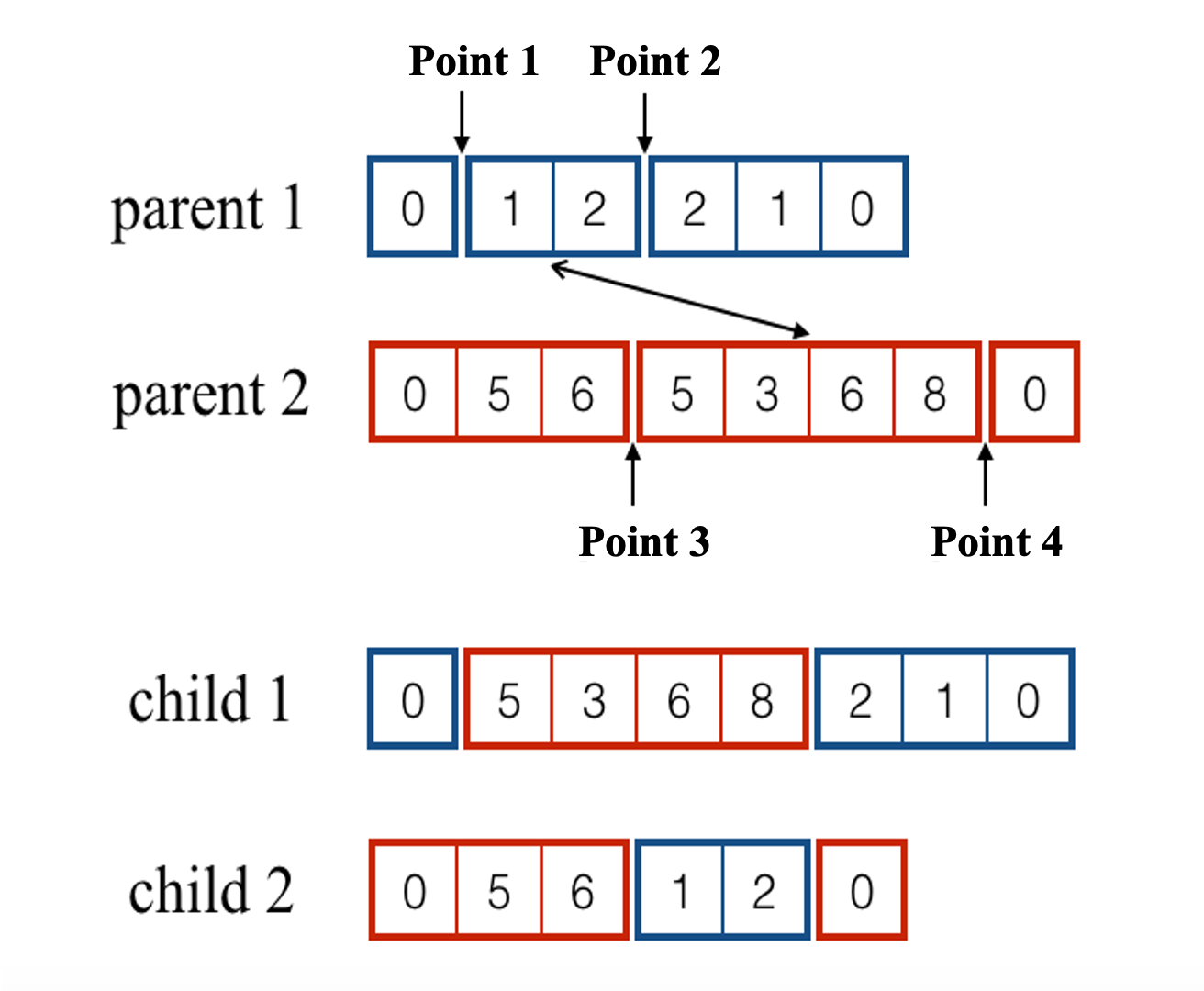}
\label{fig:crossover3}}
\caption{Examples of crossover operators in the GA}
\label{fig:crossover}
\end{figure}

We also investigate a Genetic Algorithm (GA) approach to tackle the pricing subproblem. The motivations are two-fold: first, at each column generation iteration, we need to obtain a set of routes with the most negative reduced costs, which the VNS may struggle to achieve as a single point search method. The GA is potentially more powerful as it can find a population of routes through evolution. Secondly, we believe that high quality routes (i.e. most reduced costs) may share some common structures which could be evolved more efficiently through crossover operations in the genetic algorithm. Therefore, each chromosome in our generic algorithm stands for a vehicle route, leading to a variable length chromosome. More specifically, a route(chromosome) is represented as a list of nodes(genes). For example, the parent 1 illustrated in Figure \ref{fig:crossover} simply represents route $0 \rightarrow 1 \rightarrow 2 \rightarrow 2 \rightarrow 1 \rightarrow 0$. 

The pseudo-code for the GA search is given in algorithm \ref{algo:ga}. Similar to our VNS implementation, the initial population is generated by using the insertion heuristic by \cite{chen2013task}. The size of the initial population for each RMP iteration is equal to the number of distinct vehicle routes used in the solution $\mathbf{z}$ but increased to a pre-defined value $populationSize$ in the subsequent generations. Other implementation details of our GA are as follows. Two-point crossover operators were adopted. The length between two crossover points is randomly generated from 0 to 2 arcs, as larger crossover length would increase the possibility of generating infeasible routes due to the violation of routes' travel time constraint. Figures \ref{fig:crossover1}, \ref{fig:crossover2} and \ref{fig:crossover3} illustrate examples of the two-point crossover. A standard mutation operator is used in which each chromosome is subject to an uniform \textit{2-opt} mutation with probability $mutationRate$. The 2-opt mutation operator is the same as the 2-opt neighbourhood moves in our VNS method. A local search stage is incorporated into our GA to ensure that local optima are reached in each generation. The local search is performed every time when new individuals have been generated. More specifically, the local search phase swaps two nodes between two different routes and returns two new routes that are local optimal with regard to the neighbourhood. 

We use the tournament selection method. As the first population is obtained by the insertion heuristic, it usually has smaller population size than the predetermined constant value ($populationSize$). The tournament size is set to $populationSize\times tournamentRate$ so that it is population dependent. The fitnesses of individuals are calculated according to the functions in Section \ref{pricing-methods}. Note that only feasible routes that satisfy the time constraints are considered and evaluated. If their fitnesses are better than any of the routes in the $columnPool$ which stores the set of best routes so far, they replace the inferior routes in the $columnPool$, to allow a maximum of $maxColumns$ columns to be stored. Finally, the algorithm terminates when the number of RMP iterations reaches a predefined parameter,$maxIterations$. The pseudo-code of the proposed GA is given in Algorithm \ref{algo:ga}.

Note that although the main framework of our GA is the same as many other GA implementations, the goal is very different. Our GA here does not solve the overall problem, but rather evolves a set of vehicle routes (columns) with the most negative reduced costs. These set of routes will then be used in solving the updated RMP problems. 

\begin{algorithm}[htbp]
\small
\caption{Pseudo-code of the GA column generator}
\label{algo:ga}
{\fontsize{11}{12}\selectfont
\begin{algorithmic}[1]
\Require $maxIterations$, $\mathbf{z}$, $generations$, $populationSize$, $columnPool$, $maxColumns$
\While{$i<maxIterations$}
\State $R\leftarrow \mathbf{z}$, Clear $columnPool$
\While{$j<generations$}
\State $R\leftarrow generateNewPolulation(\mathbf{z},R,populationSize)$ \Comment{Algorithm\ref{algo:ga_new_pop}}
\State $j\leftarrow j+1$
\EndWhile
\State $\mathbf{z}\leftarrow RMP(columnPool)$
\State $i\leftarrow i+1$
\EndWhile\\
\Return $\mathbf{z}$
\end{algorithmic}
}
\end{algorithm}

\begin{algorithm}[htbp]
\small
\caption{Pseudo-Code of $generateNewPolulation()$}
\label{algo:ga_new_pop}
{\fontsize{11}{12}\selectfont
\begin{algorithmic}[1]
\Require current solution $\mathbf{z}$, current population $R$, empty new population $R'$, $populationSize$, $tournamentRate$, $mutationRate$
\For {$r$ in $\mathbf{z}$} 
\If{$r$ not in $R$}
\State $R.add(r)$ \Comment{Ensure $R$ include $\mathbf{z}$}
\EndIf
\EndFor
\\
\While{$R'.size<populationSize$}
\State $r_1\leftarrow$ tournamentSelection($populationSize, tournamentRate, R$)
\State $r_2\leftarrow$ tournamentSelection($populationSize, tournamentRate, R$)
\State $R''\leftarrow$ crossover($r_1$,$r_2$)
\State $R''\leftarrow$ mutation($R''$,$mutationRate$)
\State $R''\leftarrow$ localSearch($R''$)
\For{$r$ in $R''$}
\If{fitness($r$)$<$0}
\State $R'$.add($r$)
\State updateColumnPool($r$)
\EndIf
\EndFor
\EndWhile\\
\Return $R'$
\end{algorithmic}
}
\end{algorithm}

\section{Experiments with small $R$}\label{sec:small_R}
For the first round of experiments, we consider instances with relatively small $R$. As such, all instances in the first round of experiments have seven nodes, resulting in a total of 61365 feasible routes which is close to the limit to which our model can be solved directly. Therefore, we can compare how our methods perform in comparison with exact methods. 

A set of randomly generated instances are used in the experiments. These instances are generated based on characteristics of real-life instances which are obtained from historically scheduled container operation data of a truck company. All artificially generated instances have three planning horizons of 4, 6, 8, reflecting the different problem scenarios in practice. These instances were grouped into three sets. All the instances are generated by the same parameters except the size of the planning horizon. Five instances are generated for each problem set, referred to as I4, I6 and I8, standing for shift length of 4, 6 and 8, respectively. The information and configuration of these problem sets is illustrated in Tables \ref{tab:configuration} and \ref{tab:artificial_instances}.

\begin{table}[htbp]
\small
 \centering
 \caption{Configuration of the artificial instances}
 \begin{tabular}{rl}
 \hline
 no. of nodes: & 7 (including the depot) \\
 Commodity Time Window: & 1-2 hours up to the length of planning horizon \\
 Commodity Availability Time: & nearly 30\% commodities are available at the start of\\
 & the planning horizon \\
 Emergency tasks: & 10\% to 30\% of total commodities (i.e. time window$<$10h)\\
 \hline
 \end{tabular}%
 \label{tab:configuration}%
\end{table}%

\begin{table}[htbp]
\small
 \centering
 \caption{Details of the artificial instances}
 \begin{tabular}{|c|c|c|c|}
 \hline
 Instance & no. of shifts & no. of commodities & no. of FTL units \\
 \hline
 I4-1 & 4 & 51 & 360 \\
 I4-2 & 4 & 56 & 340 \\
 I4-3 & 4 & 50 & 266 \\
 I4-4 & 4 & 87 & 624 \\
 I4-5 & 4 & 71 & 305 \\ \hline
 I6-1 & 6 & 77 & 489 \\
 I6-2 & 6 & 79 & 564 \\
 I6-3 & 6 & 94 & 581 \\
 I6-4 & 6 & 105 & 783 \\
 I6-5 & 6 & 99 & 818 \\ \hline
 I8-1 & 8 & 106 & 888 \\
 I8-2 & 8 & 120 & 831 \\
 I8-3 & 8 & 106 & 939 \\
 I8-4 & 8 & 124 & 1067 \\
 I8-5 & 8 & 127 & 971 \\
 \hline
 \end{tabular}%
 \label{tab:artificial_instances}%
\end{table}%

In order to test the efficiency of the column generation process in the first round of experiments, the initial route set is constructed by the simple method detailed in section \ref{ini_routes_basic}. Since the RMP solving will take the majority of computational time, at each iteration, we add multiple columns in the RMP model (capped by $maxColumns$). If $maxColumns$ is set too small, more RMP solving calls are required which are computationally very expensive. However, if the $maxColumns$ is set too large, time to solve each RMP would also increase (the extreme case is that all feasible columns are included in RMP and it is equivalent to the original problem). Some initial experiments suggest that $maxColumns$ = 1000 provides a good trade-off. We use this value on the understanding that it may not be the best parameter for every instance. Note that in our method, in the early search stage, we permit our method to use more trucks than the limit ($n$), but this constraint will later be restored at the end of the column generation procedure. 
Gurobi 8 linear programming libraries were used in conjunction with Java 7.0. These experiments were run on a PC with an Intel i7 3.40GHZ processor and 16GB RAM. 

The experimental results are given in Table \ref{tab:comp_two_pricing_artificial7nodes}. Since the pricing by enumeration method (see Section \ref{sec:enumeration}) takes an unrealistically long time even for the smallest instances (e.g. 3-4 hours for a 4-shift instance), it is not used for further experiments. Column \textbf{T} is the total running time of the entire process, from data parsing, solving, to the solution output. \textbf{Col.} shows the total number of columns being generated during the process. \textbf{Obj.} gives the objective value which is the total travel distance. Hereafter, \textbf{P1} and \textbf{P2} are short abbreviations for column generation solution methods adopting \textbf{P1 average pricing} and \textbf{P2 demand weighted average} respectively (see Section \ref{pricing-methods}). 

Overall, the results in Table \ref{tab:comp_two_pricing_artificial7nodes} show that most instances are solved in 1000s or less. In most cases, P2 generated a larger number of columns than P1 during the column generation process. On average, P2 generates 1165 more columns than P1, resulting in longer running times, but P2 uses 3089km less distance than P1. Seemingly, this fact is due to P2 generating more columns that enlarge the search space used by the model. However, we notice that for the result of instances I4-1, I4-2 , I8-1 and I8-3, P1 obtained a larger number of columns which did not result in a smaller objective value.


\begin{table}[htbp]
\small
 \centering
 \caption{Comparison of the two pricing methods (artificial data)}
\makebox[\linewidth]{
 \begin{tabular}{|c|c|c|c|c|c|c|c|c|c|c|c|}
 \hline
 & \multicolumn{3}{c}{P1} & \multicolumn{3}{|c|}{P2} & \multicolumn{3}{c}{Gurobi IP Solver} & \multicolumn{1}{|c|}{Random} \\ \cline{2-4} \cline{5-7} \cline{8-10} \cline{11-11}
 Instance & T & Obj. & Col. & T & Obj. & Col. & T & Obj. &Obj.*& Obj. \\ \hline
 I4-1 & 23 & 15516 & 4691 & 21 & 14154 & 3005 & 1215 & 13746 & n.a. & 22530 \\
 I4-2 & 93 & 16480 & 9448 & 151 & 15988 & 5976 & 1208 & 15823 & n.a.& 22743 \\
 I4-3 & 3 & 12793 & 2281 & 6 & 11067 & 4319 & 155 & 11037 & n.a.& 15885 \\
 I4-4 & 271 & 27557 & 4674 & 711 & 25642 & 8811 & 1736 & 25307 & 28819& 33583 \\
 I4-5 & 187 & 13407 & 4021 & 343 & 11435 & 6430 & 1193 & 11429 & 14624& 25798 \\
 I6-1 & 131 & 27566 & 7742 & 193 & 25540 & 13985 & 1153 & 24713 & 29542& 34589 \\
 I6-2 & 175 & 26719 & 3046 & 507 & 23374 & 4861 & 2772 & 21665 & n.a.& 31294 \\
 I6-3 & 87 & 32009 & 3142 & 513 & 30124 & 5321 & 2604 & 30029 & n.a.& 31889 \\
 I6-4 & 218 & 41301 & 2170 & 290 & 35935 & 3321 & 9462 & 33898 & n.a.& 54497 \\
 I6-5 & 172 & 33799 & 2040 & 420 & 30207 & 3216 & 5406 & 29223 & n.a.& n.a. \\
 I8-1 & 276 & 53871 & 2863 & 694 & 50178 & 2724 & 14890 & 49797 & n.a.& 70269 \\
 I8-2 & 323 & 38589 & 2199 & 958 & 33532 & 3361 & 17202 & 32668 & n.a.& 54667 \\
 I8-3 & 213 & 44856 & 3539 & 970 & 39643 & 2701 & 10006 & 38108 & n.a.& n.a. \\
 I8-4 & 479 & 35850 & 2022 & 919 & 32307 & 2286 & 36132 & 31979 & n.a.& 46778 \\
 I8-5 & 213 & 45066 & 2414 & 704 & 39911 & 3455 & 19170 & 37979 & n.a.& 61476 \\ \hline
 Avg. & 191 & 31025 & 3753 & 493 & 27936 & 4918 & 8287 & 27160 & n.a.& 38923 \\
 \hline
\multicolumn{11}{l}{P:pricing method; T:Total running time(s);}\\
\multicolumn{11}{l}{Col.:Total columns generated; Obj.:Objective value(km).}\\
\multicolumn{11}{l}{Obj*.:Objective value with a limited computational time.}\\
\multicolumn{11}{l}{n.a.:Failed to find feasible solution in the given time.}\\
 \end{tabular}%
 \label{tab:comp_two_pricing_artificial7nodes}%
}
\end{table}%

The performance of both algorithms is also compared with the results from the Gurobi IP solver with the default algorithm setting in two experiments. The first experiment allows the solver to solve the problem to optimality and its objective value is denoted as \textit{Obj.}. In the second experiment, Gurobi was given a limited computational time (the same time taken by the slowest of P1 and P2) and the corresponding objective value is marked as \textit{obj.*}. All the results are given in Table \ref{tab:comp_two_pricing_artificial7nodes}. 

It can be seen that although Gurobi can solve all instances to optimality, it takes more than 8000s on average and sometimes more than 10h. Two tailed paired t-tests ($\alpha$ = 0.01) were conducted to compare the performance between P1, P2 and Gurobi. In contrast, the proposed column generation methods (P1 and P2) use significantly less time (P1 vs. Gurobi: t=-3.389, p$<$0.01; P2 vs. Gurobi: t=-3.298, p$<$0.01) with competitive solutions. This is particularly true when P2 is used as, on average, it uses around 5\% of the time used by Gurobi but produces solutions that are only 776km (or 2.80\%, P2 vs. Gurobi: t=4.517, p$<$0.01) away from optimality. On the other hand, if we reduce computational time, for many instances Gurobi fails to produce a feasible solution. Between P1 and P2, P1 generates less columns and is faster, but produces inferior solutions for most instances. 

To evaluate the usefulness of the pricing subproblem, we conducted another set of experiments by replacing the routes produced by the pricing subproblem with $maxColumns$ randomly selected routes (from all possible feasible routes) of the RMP at each iteration. All the other settings were kept the same as before. Column \textbf{Random} in Table \ref{tab:comp_two_pricing_artificial7nodes} presents the objective values based on the average of five runs. As can be seen, the results are significantly inferior to those by P1 or P2 (P1 vs. Random: t=-6.496, p$<$0.01; P2 vs. Random: t=-6.993, p$<$0.01), which shows the importance of the pricing subproblem.

\section{Experiments on instances with a very large $R$}\label{sec:large_R}
For larger instances, the feasible route set $R$ can become very big and therefore it becomes impossible to enumerate them all as we did in the previous section. In this section, we investigate the effectiveness and performance of the two metaheuristic approaches presented in section \ref{sec:heuristic-pricing}. As the evidence from previous experiments suggest P2 performs better than P1, for the remaining experiments only P2 is used. Similar to the previous section, maximum $maxColumns=1000$ columns are allowed to be generated by both VNS and GA at each iteration. As $maxColumns$ is the only parameter used in the proposed VNS based column generator, parameter tuning for VNS is omitted. We now illustrate parameter tuning for the GA.

\subsection{Parameter Tuning for GA}\label{sec:ga_parameters}
The parameters used in the proposed GA are the population size (\textit{populationSize}), the generation size (\textit{generations}), the probability of mutation (\textit{mutationRate}), and the tournament size i.e. the tournament rate (\textit{tournamentRate}). In this experiment, the \textit{mutationRate} is set to 0.02 and the \textit{tournamentRate} is set to 0.1 after some initial tuning. Table \ref{tab:test_p_g} shows the results with the algorithm with different population sizes and different number of generations for the two most challenging problem instances LB8-1 and LB8-2. Each instance was run five times and the average result of both instances is given in column\textit{ Avg.}. The $maxIteration$ is set to 5 as increasing it further gives very little further improvement. With the consideration of algorithm efficiency, we choose the combination of \textit{populationSize}=500 and \textit{generations}=500.

\begin{table}[htbp]
\small
  \centering
 \caption{Experiment results for evaluating \textit{populationSize} and \textit{generations}}
    \begin{tabular}{|c|c|c|c|c|c|c|c|c|c|}
    \hline
    populationSize & 10    & 100   & 200   & 500   & 10    & 100   & 200   & 500   & 10 \\
    generations & 10    & 10    & 10    & 10    & 100   & 100   & 100   & 100   & 200 \\
    Avg.  & 25043 & 22154 & 21797 & 21280 & 21930 & 20991 & 20588 & 20079 & 21647 \\
\hline
    populationSize & 100   & 200   & 500   & 10    & 100   & 200   & 500   & 1000  & 2000 \\
    generations & 200   & 200   & 200   & 500   & 500   & 500   & 500   & 1000  & 2000 \\
    Avg.  & 20676 & 20128 & 19999 & 21205 & 20228 & 20081 & 19879 & 19775 & 19724 \\
   \hline
    \end{tabular}%
 \label{tab:test_p_g}%
\end{table}%

\subsection{Comparing VNS and GA based column generation approaches}
Due to the very large amount of computational time required, we select a total of six instances, two from the real-life instance set and four from the artificial instance set used by \cite{Bai2015134}. The instance names starting with \textbf{NP} are real-life instances while those starting with \textbf{LB}, \textbf{TB}, \textbf{LU} and \textbf{TU} represent (\textbf{Loose, Balanced}), (\textbf{Tight, Balanced}), (\textbf{Loose, Unbalanced}) and (\textbf{Tight, Unbalanced}) configurations of artificial instances. The first digit of each instance name indicates the length of the planning horizon (e.g. NP4-1 is a real-life instance with a 4-shift planning horizon). 

\begin{figure}[htbp]
\centering
\subfloat[][Instance NP4-1]{
\includegraphics[width=0.4\textwidth]{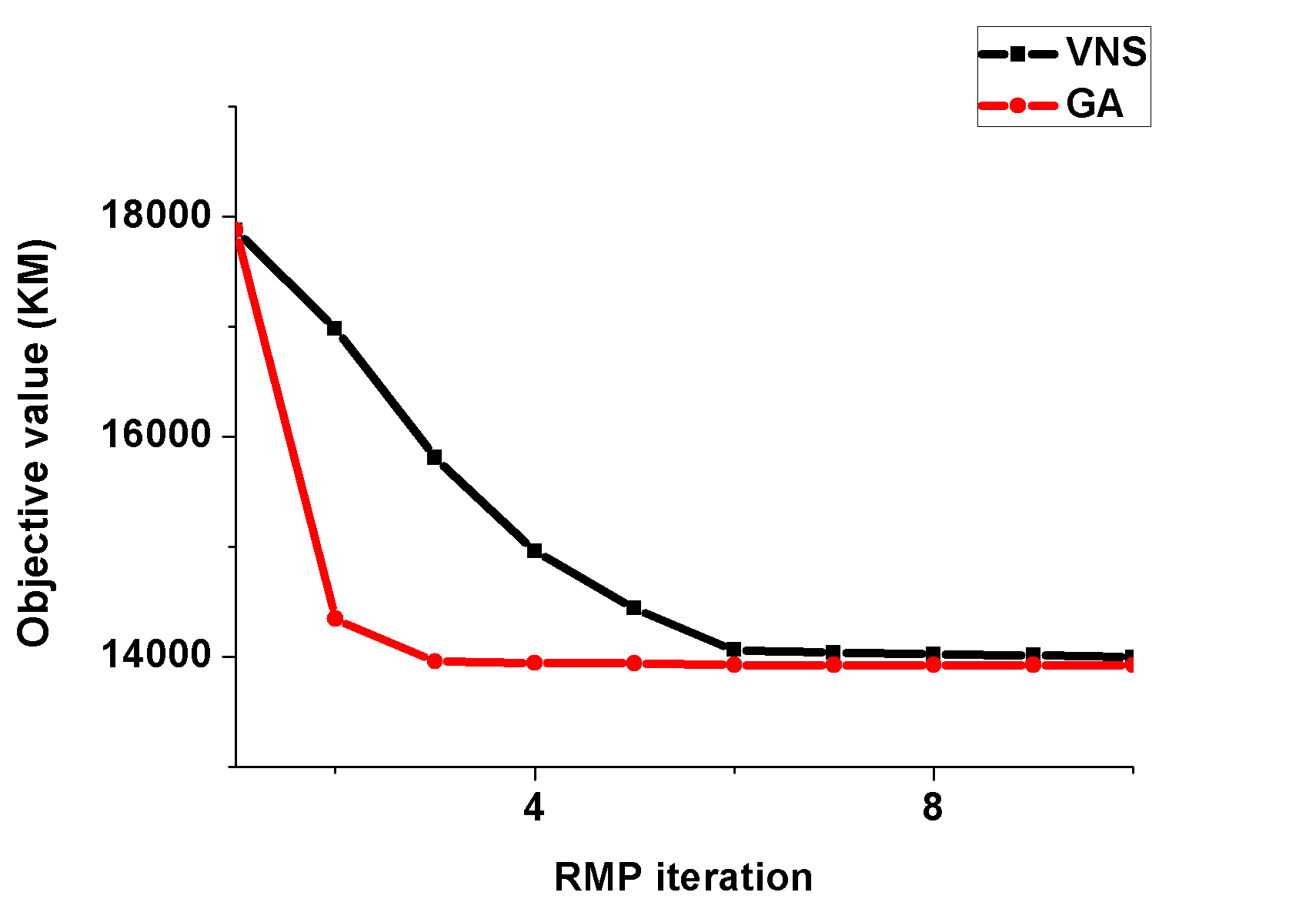}
\label{fig:np4-1}}
\subfloat[][Instance NP8-1]{
\includegraphics[width=0.4\textwidth]{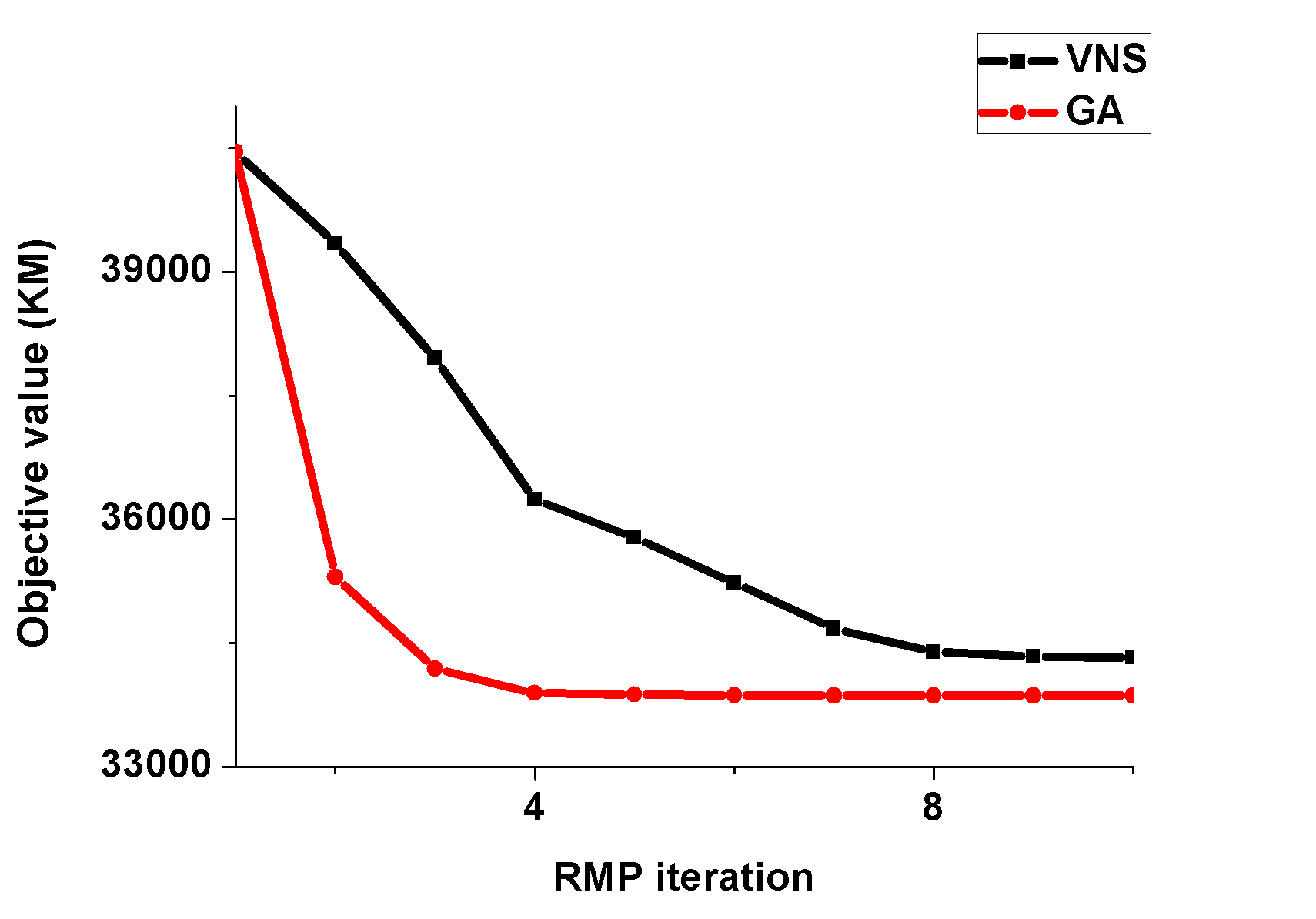}
\label{fig:np8-1}}
\qquad
\subfloat[][Instance LB8-1]{
\includegraphics[width=0.4\textwidth]{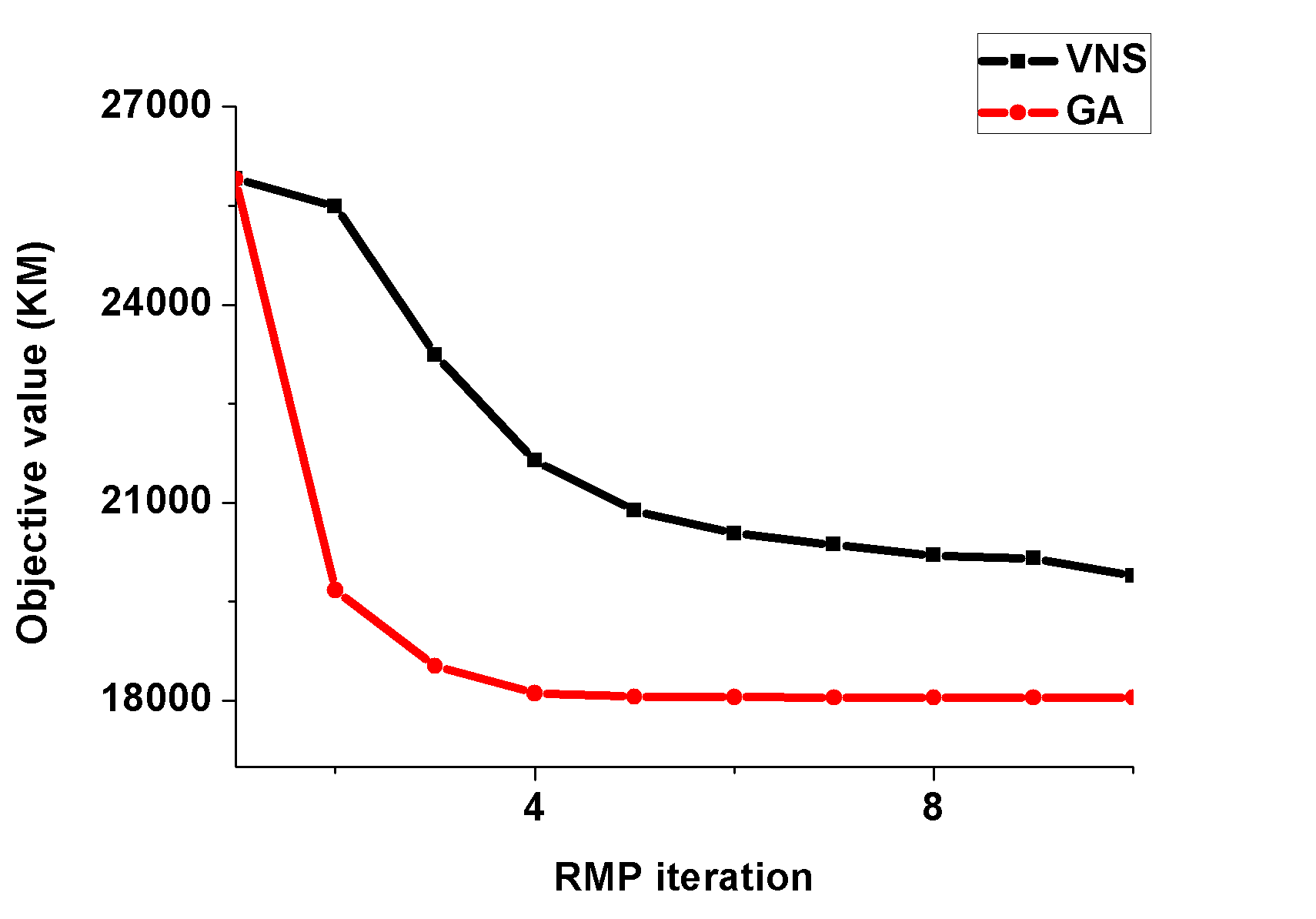}
\label{fig:lb8-1}}
\subfloat[][Instance TU8-7]{
\includegraphics[width=0.4\textwidth]{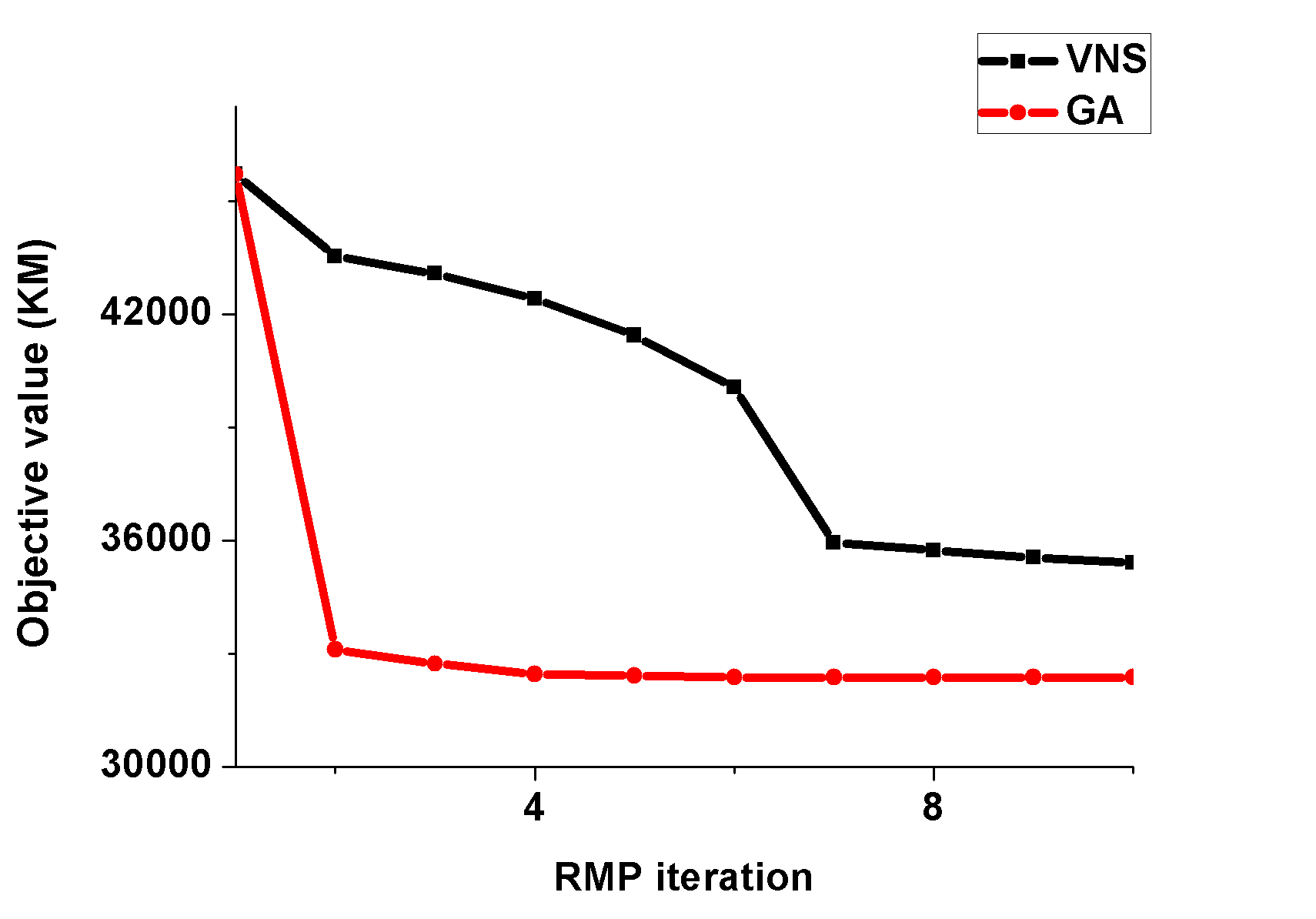}
\label{fig:tu8-7}}
\qquad
\subfloat[][Instance TB4-4]{
\includegraphics[width=0.4\textwidth]{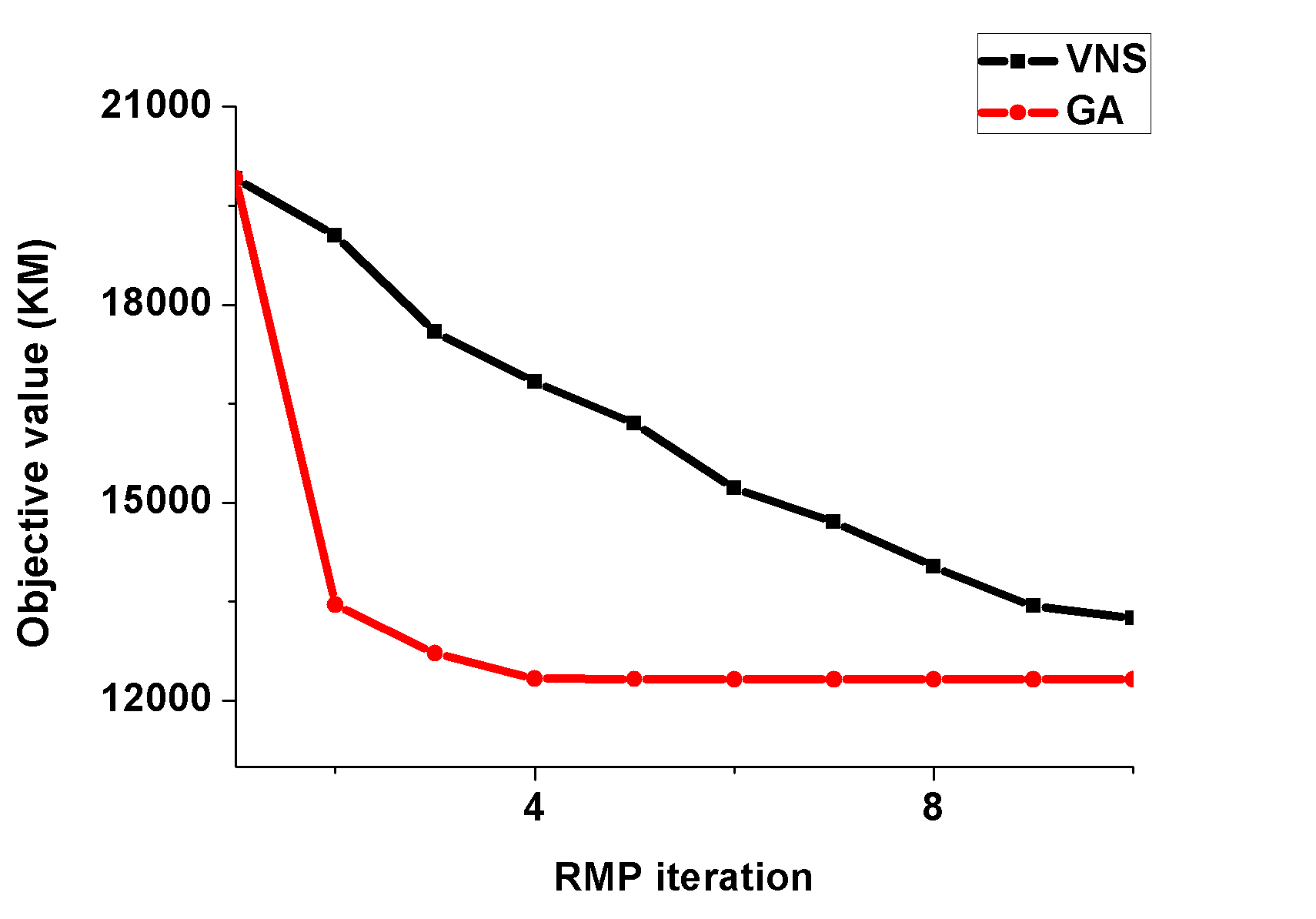}
\label{fig:tb4-4}}
\subfloat[][Instance LU4-6]{
\includegraphics[width=0.4\textwidth]{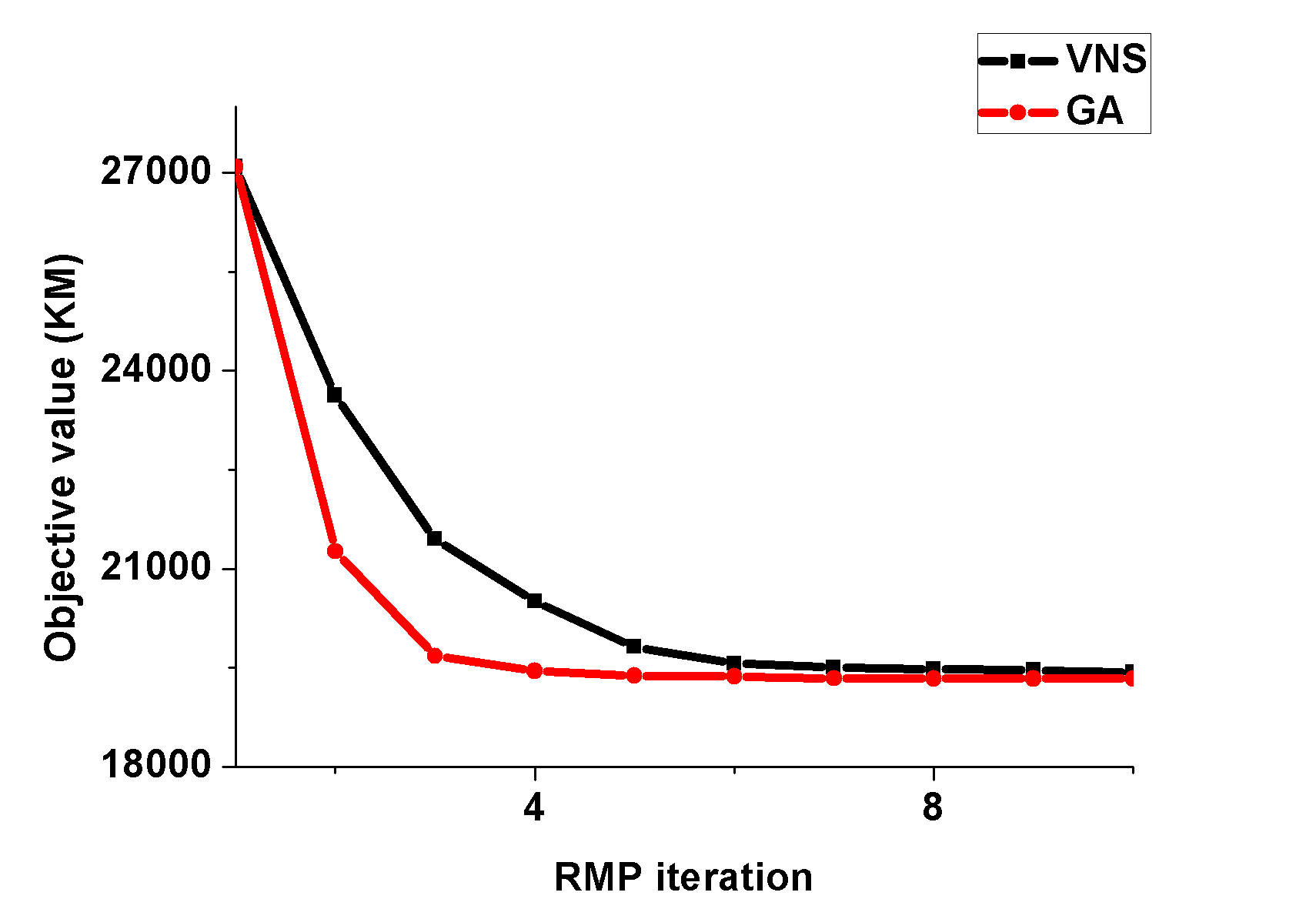}
\label{fig:lu4-6}}
\caption{Comparison of performance \textit{maxIteration}=10.}
\label{fig:vns_ga_compare}
\end{figure}

The stopping criterion \textit{maxIteration} is set to 10 for both VNS and GA. Figure \ref{fig:vns_ga_compare} lists the average objective values of five runs of experiments. The horizontal axis defines the number of RMP iterations while the vertical axis given the objective values. The results suggest that the GA is a faster converging algorithm thanks to its population based search framework and capacity to evolve a set of routes instead of a single one. 

As experiments show that the VNS converges in 30 RMP iterations, for a fairer comparison, we set \textit{maxIteration} to 30 for both GA and VNS for all instances. In addition, a comparison was also made with previous results obtained by the three-stage method, meta-heuristic methods and lower bound reported in \cite{Bai2015134}. 

Table \ref{tab:compare_all} presents the running time and objective values obtained by the pricing method P2 using VNS and GA column generators (\textit{P2-VNS}, \textit{P2-GA}), \textit{3-stage} method \citep{Bai2015134}, meta-heuristic methods using simulated annealing hyper-heuristic (\textit{SAHH}), and reactive shaking variable neighbourhood search (\textit{rsVNS}). In terms of objective values, the average results in increasing order are as follows: \textit{P2-GA} $<$ \textit{3-stage} $<$ \textit{P2-VNS} $<$ \textit{SAHH} $<$ \textit{rsVNS}. The best results are highlighted in bold. The average running times show an increasing order as follows: \textit{P2-VNS} $<$ \textit{SAHH} $<$ \textit{P2-GA} $<$ \textit{rsVNS} $<$ \textit{3-stage}. Two tailed Paired t-tests ($\alpha$ = 0.01) were conducted to compare P2-VNS and P2-GA with 3-stage, SAHH and rsVNS approaches. There are significant difference in the solution quality for P2-VNS vs. P2-GA(t=3.698, p$<$0.01), P2-VNS vs. rsVNS(t=-4.239, p$<$0.01), P2-VNS vs. SAHH(t=-3.916, p$<$0.01), P2-GA vs. rsVNS(t=-5.667, p$<$0.01), P2-GA vs. SAHH(t=-5.463, p$<$0.01). These tests suggest that both \textit{P2-GA} and \textit{P2-VNS} are able to find better solution in less time than most of the existing algorithms. 


The novel solution coding and pricing methods limit the search space for the algorithm, so its efficiency is increased compared with the results obtained by meta-heuristics (\textit{rsVNS} and \textit{SAHH}). The \textit{3-stage} method performs well for tight instances, but it does less well for large and loose instances. The reason is that it employs an integer programming solver so its solution time increases exponentially with large problem sizes. The proposed column generation methods are able to find effective columns in order to reduce the problem size, therefore, compared with the \textit{3-stage} method, the solving time of column generation method is significantly decreased for large instances. However, the advantage of column generation method may not be obvious for small problem instances (i.e. tight and small instances) as the iterative RMP solving comprises a significant proportions of the run time by the algorithm. 

\begin{table}[htbp]
\small
 \centering
 \caption{Comparisons with previous results}
\makebox[\linewidth]{
 \begin{tabular}{|c|c|c|c|c|c|c|c|c|c|c|}
 \hline
 & \multicolumn{2}{c|}{\textit{P2-VNS}} & \multicolumn{2}{c}{\textit{P2-GA}} & \multicolumn{2}{|c|}{\textit{3-stage}} & \multicolumn{2}{c}{\textit{rsVNS}} & \multicolumn{2}{|c|}{\textit{SAHH}} \\
 \hline
 Instance & T & Obj. & T & Obj. & T & Obj. & T & Obj. & T & Obj. \\ \hline
 NP4-1 & 126 & 13978 & 405 & 13860 & 33301 & \textbf{13509} & 4800 & 14453 & 310 & 14471 \\
 NP4-2 & 112 & 16667 & 462 & \textbf{16621} & 15742 & 16636 & 4800 & 16593 & 386 & 16595 \\
 NP4-3 & 132 & 17110 & 417 & 17106 & 11178 & \textbf{16879} & 4800 & 17138 & 590 & 17383 \\
 NP4-4 & 273 & 22100 & 509 & \textbf{21980} & 18537 & 21886 & 4800 & 22302 & 545 & 22142 \\
 NP4-5 & 384 & 26184 & 1195 & \textbf{26166} & 20647 & 26731 & 4800 & 26216 & 1022 & 26239 \\
 NP6-1 & 1017 & 34054 & 3742 & \textbf{34022} & 160079 & 34055 & 7200 & 35209 & 1613 & 35122 \\
 NP6-2 & 1360 & 33490 & 1868 & 33490 & 138486 & \textbf{33316} & 7200 & 33808 & 955 & 33653 \\
 NP6-3 & 45 & 16150 & 198 & \textbf{16094} & 3978 & 16192 & 7200 & 16660 & 211 & 16247 \\
 NP6-4 & 356 & 26146 & 1262 & \textbf{26126} & 58898 & 26260 & 7200 & 26272 & 698 & 26316 \\
 NP6-5 & 545 & 16883 & 984 & \textbf{16817} & 104446 & 16881 & 7200 & 17950 & 492 & 17800 \\
 NP8-1 & 730 & 33889 & 1133 & \textbf{33789} & 148067 & 35685 & 9600 & 34181 & 822 & 34095 \\
 NP8-2 & 825 & 30576 & 1612 & \textbf{30554} & 147241 & 30633 & 9600 & 31639 & 869 & 31310 \\
 NP8-3 & 1049 & 28281 & 1260 & \textbf{28281} & 121074 & 28314 & 9600 & 28450 & 878 & 28451 \\
 NP8-4 & 1211 & 43643 & 1731 & \textbf{43630} & 66438 & 44224 & 9600 & 43955 & 1631 & 43943 \\
 NP8-5 & 898 & 25419 & 1415 & \textbf{25389} & 131369 & 25452 & 9600 & 25742 & 1128 & 26182 \\
 LB4-1 & 89 & 15852 & 447 & 15766 & 13438 & \textbf{15763} & 4800 & 16011 & 292 & 15865 \\
 LB4-2 & 61 & 14975 & 283 & 14777 & 3812 & \textbf{14319} & 4800 & 15291 & 414 & 15059 \\
 TB4-3 & 30 & 11027 & 128 & \textbf{10364} & 1415 & 10867 & 4800 & 11027 & 288 & 11092 \\
 TB4-4 & 21 & 12671 & 157 & \textbf{12172} & 186 & 12508 & 4800 & 13577 & 383 & 13495 \\
 LU4-5 & 66 & 18242 & 183 & \textbf{17676} & 1590 & 18500 & 4800 & 19884 & 276 & 19717 \\
 LU4-6 & 65 & 19403 & 215 & \textbf{19394} & 1783 & 20316 & 4800 & 19741 & 233 & 19859 \\
 TU4-7 & 6 & 12869 & 113 & \textbf{12804} & 79 & 13033 & 4800 & 13760 & 232 & 14377 \\
 TU4-8 & 12 & 18920 & 125 & 17956 & 138 & \textbf{17025} & 4800 & 17846 & 491 & 17815 \\
 LB8-1 & 375 & 18251 & 1803 & \textbf{18097} & 138988 & 18133 & 9600 & 18542 & 1899 & 18325 \\
 LB8-2 & 444 & 22265 & 2909 & \textbf{20928} & 157354 & 22834 & 9600 & 23068 & 1266 & 22990 \\
 TB8-3 & 73 & 21670 & 224 & \textbf{20456} & 148 & 21338 & 9600 & 21657 & 2602 & 21689 \\
 TB8-4 & 112 & 28001 & 193 & \textbf{25316} & 561 & 28167 & 9600 & 28398 & 2391 & 28305 \\
 LU8-5 & 73 & 23288 & 248 & 22453 & 4380 & \textbf{21226} & 9600 & 24587 & 915 & 24787 \\
 LU8-6 & 226 & 23528 & 659 & \textbf{22690} & 13202 & 23261 & 9600 & 24412 & 1204 & 24261 \\
 TU8-7 & 58 & 32680 & 166 & 32334 & 140 & \textbf{31094} & 9600 & 35595 & 484 & 35581 \\
 TU8-8 & 58 & 27884 & 197 & \textbf{26958} & 66 & 27406 & 9600 & 28197 & 434 & 28162 \\
 Large & 8355 & 105793 & 7801 & \textbf{100119} & n.a. & n.a. & 9600 & 142258 & 15848 & 141252 \\ \hline
 Average & 349 & 22777 & 847 & \textbf{22389} & 48928 & 22659 & 7200 & 23295 & 837 & 23269 \\
 \hline
 \multicolumn{11}{l}{n.a.:Failed to find feasible solution in given time.}\\
\multicolumn{11}{l}{Average: ignoring the results of the \textbf{Large} instance.}\\
 \end{tabular}%
 \label{tab:compare_all}%
}
\end{table}%

\section{Conclusions}\label{sec:conclusion}
We have presented an innovative FTL routing formulation assisted by dynamic cuts and investigated column generation based approaches which are particularly effective on very large instances. Unlike traditional branch-price-and-cut, it performs an incomplete search, with the aim of finding good solutions more quickly. It efficiently solves the problem using the following strategies: 1) Infeasible flow assignments are allowed in the column generation process but will be fixed by adding cuts in the end; 2) To reduce the number of decision variables, some constraints are pre-processed offline; 3) The shift that a route belongs to is not restricted;  4) To avoid traversing the entire search tree,  metaheuristics are implemented to repeatedly identify new columns with negative reduced costs in light of both new pricing information and latest columns on the basis of the RMP problem.

Two pricing methods and two approaches for generating initial routes were proposed and evaluated. The result indicates that the proposed solution methods improve the existing algorithms both in terms of the computational time and the solution quality. We believe the advantageous features of the indirect solution encoding of the FTL problem haven been fully explored in this paper and the proposed solution methods can efficiently solve real-life drayage container operation problem with long planning horizon covering multi-shifts.

\section*{Acknowledgments}
This work is supported by the National Natural Science Foundation of China (Grant No. 71471092),
Natural Science Foundation of Zhejiang Province (Grant No. LR17G010001) and Ningbo Municipal Bureau of Science and Technology (Grant No. 2017D10034).




\bibliographystyle{elsarticle-harv}
\bibliography{Optimization_bib}{}
\end{document}